\documentclass[preprint, 10pt]{elsarticle}
\usepackage[margin=2.5cm]{geometry}
\usepackage{setspace}
\usepackage{lmodern}
\usepackage{amsmath,amssymb}
\usepackage{amsfonts,amsthm}
\usepackage{lineno}
\usepackage{graphicx}
\usepackage{bm, bbm}
\usepackage{algorithm}
\usepackage{algorithmic}
\usepackage{physics}
\usepackage{color}
\usepackage{pxfonts}
\usepackage[footnotesize,bf]{caption}
\usepackage{subcaption}
\usepackage{mathtools}
\usepackage{hhline}
\usepackage[T1]{fontenc}
\usepackage{multirow}
\usepackage{placeins}
\usepackage{float}
\usepackage{tabularx}
\usepackage{hyperref}
\usepackage{xcolor}
\usepackage{makecell}
\hypersetup{
	colorlinks=false,
	pdfborder={0 0 1},
	linkbordercolor={0 1 0},
	citebordercolor={0 1 0},
	filebordercolor={0 1 0},
	urlbordercolor={0 1 0}
}

\newtheorem{remark}{Remark}[section]

\renewcommand{\arraystretch}{1.5}
\graphicspath{{}}
\numberwithin{equation}{section}
\usepackage{listings}
\usepackage{xcolor}
\newcommand{\ignore}[1]{}
\usepackage{graphicx} % 必须加载宏包
\newcommand{\myO}{\scalebox{0.8}{$\mathcal{O}$}}
\usepackage{booktabs}

\usepackage{xcolor}
\usepackage{etoolbox}

\begin{document}
\begin{frontmatter}
\title{Deep set based operator learning with uncertainty quantification}
\author[SNH]{Lei Ma}
\author[SNH]{Ling Guo}

\author[SDU]{Hao Wu\corref{cor}}
\cortext[cor]{Corresponding Author}\ead{hwu81@sjtu.edu.cn.}
\author[SAC]{Tao Zhou}

\address[SNH]{Department of Mathematics, Shanghai Normal University, Shanghai, China}
\address[SDU]{School of Mathematical Sciences, Institute of Natural Sciences, and MOE-LSC, Shanghai Jiao Tong University, Shanghai, China}
\address[SAC]{Institute of Computational Mathematics and Scientific/Engineering Computing, AMSS, Chinese Academy of Sciences, Beijing, China}

\begin{abstract}
Learning operators from data is central to scientific machine learning. While DeepONets are widely used for their ability to handle complex domains, they require fixed sensor numbers and locations, lack mechanisms for uncertainty quantification, and are thus limited in practical applicability. Recent permutation-invariant extensions, such as the Variable-Input Deep Operator Network, relax these sensor constraints but still rely on sufficiently dense observations and cannot capture uncertainties arising from incomplete measurements or from operators with inherent randomness. To address these challenges, we propose UQ-SONet, a permutation-invariant operator learning framework with built-in uncertainty quantification. Our model integrates a set transformer embedding to handle sparse and variable sensor locations, and employs a conditional variational autoencoder to approximate the conditional distribution of the solution operator. By minimizing the negative ELBO, UQ-SONet provides principled uncertainty estimation while maintaining predictive accuracy. Numerical experiments on deterministic and stochastic PDEs, including the Navier–Stokes equation, demonstrate the robustness and effectiveness of the proposed framework.
\end{abstract}


\end{frontmatter}

\section{Introduction}

Learning continuous operators associated with complex systems from scattered data streams has shown promising success in scientific machine learning, which focuses on modeling mappings between infinite-dimensional function spaces. A prominent example is the Fourier Neural Operator (FNO) \cite{li2020fourier}, which parameterizes the integral kernel in Fourier space. While highly effective for problems on domains that can be discretized or efficiently mapped to Cartesian grids, the applicability of FNO to more general settings remains limited \cite{lu2022comprehensive}. Another representative approach is the Deep Operator Network (DeepONet) \cite{lu2021learning}, which employs a branch network to encode sensor measurements of the input function and a trunk network to encode the output coordinates, combining their outputs to approximate the solution operator. Compared to FNOs, DeepONets are capable of learning operators on complex domains and under more general boundary conditions, and have therefore been widely adopted in diverse engineering applications \cite{lu2021learning, mao2021deepm, cai2021deepm, lin2021operator}. Nevertheless, DeepONets still face fundamental limitations, particularly in handling incomplete or variable sensor observations, which restricts their applicability in practical scenarios.

A fundamental limitation of DeepONets lies in their reliance on a fixed set of sensor locations: the number and positions of sensors must remain identical across all training and testing samples. This assumption is often unrealistic, since data collected from experiments or simulations are frequently irregular, sparse, or available at varying locations. To mitigate this issue, several recent works have introduced permutation-invariant architectures into operator learning. For example, the Mesh-Independent neural operator method was introduced in \cite{lee2022mesh}, which  represents the discretized system as set-valued data without a prior structure and construct a permutation-invariant model that properly treats both input functions and query coordinates of the solution function. OFormer \cite{li2022transformer}, a data-driven operator learning framework based on attention mechanisms, makes few assumptions on the sampling pattern of the input function or query locations, and thus exhibits strong robustness to irregular grids. The Variable-Input Deep Operator Network (VIDON) \cite{prasthofer2022variable} was proposed as a flexible extension of DeepONet. Inspired by Deep Sets \cite{zaheer2017deep, wagstaff2019limitations} as well as related frameworks on functions defined on sets and transformer-based architectures \cite{vaswani2017attention, lee2019set}, VIDON incorporates permutation invariance into the branch network through a lightweight transformer structure, enabling the model to dynamically adapt to varying numbers and positions of input sensors. In this way, VIDON substantially enhances the flexibility and robustness of operator learning compared to classical DeepONets. Despite these advances, VIDON still relies on the assumption that input observations are sufficiently dense and informative to characterize the operator. In practical scenarios, however, sensor data may be incomplete, sparse, or corrupted by noise. Under such conditions, the solution operator inherently exhibits uncertainty, which is not explicitly addressed by VIDON.

The challenge of incomplete or noisy observations naturally calls for frameworks that incorporate uncertainty quantification (UQ) into operator learning. Several approaches have recently been proposed in this direction. For example, the Bayesian neural network demonstrated strong performance across a range of task via incorporating prior knowledge of physics \cite{yang2021b, linka2022bayesian, lin2023b}. Deep ensemble-based methods try to approximate uncertainty through variability across multiple trained models \cite{lakshminarayanan2017simple, fort2019deep}. The information bottleneck-based UQ framework was proposed in \cite{guo2024ib} for UQ in function regression and operator learning tasks. More related works that address UQ under noisy or partially observed data can be found in \cite{moya2023deeponet, psaros2023uncertainty, yang2022scalable, zhu2023reliable, zou2024neuraluq} and references therein. While these methods mark an important step toward uncertainty-aware operator learning, few studies have explicitly investigated UQ in the context of permutation-invariant operator learning frameworks.

To address this gap, we propose UQ-SONet, a permutation-invariant operator learning framework with built-in UQ. Our model employs a set transformer embedding to integrate sensor locations and measurements of the input function in a permutation-invariant manner, and a conditional variational autoencoder (cVAE) to approximate the conditional distribution of the output function given sparse observations. By minimizing the negative Evidence Lower Bound (ELBO), UQ-SONet provides a principled framework for capturing uncertainties arising from incomplete and irregular input data, while preserving accuracy in operator approximation. Our main contributions can be summarized as follows:
\begin{itemize}
\item We propose UQ-SONet, a novel operator learning framework that integrates permutation invariance with principled UQ.

\item By combining a set transformer embedding with a cVAE, the proposed model allows for sparse variable sensor locations and noisy observations while maintaining accuracy in operator approximation.

\item The framework can also be extended to operators with intrinsic randomness, such as stochastic differential equations (SDEs). While the original DeepONet framework attempted to learn stochastic operators through a Karhunen–Loève (KL) expansion, it was restricted to low-dimensional random inputs. In contrast, UQ-SONet provides a more flexible approach that accommodates higher-dimensional stochastic settings.
\end{itemize}

The rest of the paper is organized as follows. Section 2 reviews the DeepONet architecture and introduces the proposed UQ-SONet model. Section 3 details the structure of UQ-SONet, including the set transformer embedding, the encoder–decoder design of the cVAE, and the loss function used for training. Section 4 presents a series of numerical experiments to demonstrate the performance of the proposed method, covering one- and two-dimensional deterministic and stochastic PDEs, as well as the Navier–Stokes equation. Section 5 concludes the paper with a summary of our findings and outlines potential directions for future research.

\section{Problem setup}

For ease of reference, we provide a consolidated notation table at the beginning of this section. Table \ref{tab:notation} summarizes the key symbols introduced in Sections 2 and 3, together with their definitions. It includes the notation for the operator learning formulation, the observations and sensors, and the main quantities used in the proposed model architecture. 

\begin{table}[!ht]
	\centering
	\small
	\caption{\textbf{Notation.} Summary of the main symbols and definitions used in Sections 2-3, including the operator learning formulation and the components of the proposed model.}
	\label{tab:notation}
	\setlength{\tabcolsep}{16pt}
	\renewcommand{\arraystretch}{1.3}
	\begin{tabular}{llcl}
		\toprule[1.2pt]
		\textbf{Symbol}   &   \textbf{Description} \\
		\midrule
		$\kappa$ / $u$  &  input /output function of the operator \\
		
		$\mathcal{G}$ / $\hat{\mathcal{G}}$  &  true operator / surrogate operator model \\
		
		$\myO=\{(x_i,\kappa(x_i))\}_{i=1}^m$  &  set of observations of the input function \\
		
		$x$ / $y$  &  coordinates of the input/output functions \\
		
		$m$ / $\mathcal{M}$  &  number of sensors / admissible set of sensor counts \\
		
		$b^{n}$ / $t^{n}$  &  outputs of the branch/trunk networks \\
		
		$\mathcal{D}=\{(\kappa_i,u_i)\}_{i=1}^N$  &  training dataset of input--output function pairs \\
		
		$\Lambda_x$ / $\Lambda_{\kappa}$  &  coordinate/value embeddings in the input embedding module \\
		
		$d_{\mathrm{emb}}$  &  embedding dimension of each sensor representation \\
		
		$w_l(\cdot)$ / $v_l(\cdot)$  &  attention weight / value projection networks for the $l$-th head \\
		
		$H$  &  number of attention heads in the attention pooling module \\
		
		$h(\myO)$  &  permutation-invariant representation of the input observations \\
		
		$z$ / $d_z$  &  latent variable / dimension of the latent space \\
		
		$q_E$ / $p_D$  &  conditional distributions induced by the encoder/decoder\\
		
		$\mu_z$ / $\Sigma_z$  &  encoder outputs, namely the approximate posterior mean / covariance \\
		
		$\bar u=(u(y_1),\ldots,u(y_M))^\top$  &  finite-dimensional  representation of $u$ on the grid $\{y_j\}_{j=1}^M$ \\
		
		$\sigma_u^2$  &  artificial noise variance in $p_{D}$\\
		\bottomrule[1.2pt]
	\end{tabular}
\end{table}

We begin by considering the estimation of the operator \( u = \mathcal{G}(\kappa) \), where both \( u \) and \( \kappa \) have domains contained within finite-dimensional Euclidean spaces, and their ranges are subsets of \(\mathbb{R}\). DeepONets \cite{lu2021learning} have emerged as a significant machine learning approach to tackle this problem. The general structure of DeepONet comprises two primary components: the branch network and the trunk network. The branch network processes the input function \(\kappa\), which is characterized by a series of observations \( \myO = \{(x_i, \kappa(x_i))\}_{i=1}^m \) in its domain, while the trunk network processes the spatial coordinates. These two networks are then combined to produce the output function. Mathematically, for a given input \(\kappa\) and a coordinate \(y\) in the domain of \(u\), the output of DeepONet can be expressed as:
\begin{eqnarray}
\hat{\mathcal{G}}(\kappa)(y) & = & \sum_{n=1}^{p}\underbrace{b^{n}(\myO)}_{\text{branch}}\underbrace{t^{n}(y)}_{\text{trunk}}\nonumber \\
 & = & \sum_{n=1}^{p}b^{n}\left(\kappa(x_{1}),\ldots,\kappa(x_{m})\right)t^{n}(y),\label{eq:deeponet-conventional}
\end{eqnarray}
where \((b^1,\ldots,b^p)^\top\) and \((t^1,\ldots,t^p)^\top\) are neural networks with \(p\) outputs. Given a training dataset \(\mathcal{D} = \{(\kappa_i, u_i)\}_{i=1}^{N}\), all parameters of DeepONet can be optimized by minimizing the error between the predicted outputs \(\hat{\mathcal{G}}(\kappa_i)\) and the actual outputs \(u_i\). It is important to note that in the classical DeepONet framework, the sensor locations \(\{x_i\}_{i=1}^m\) in \(\myO\) are fixed during both the training and testing phases. As a result, the input to the branch network consists solely of the observed values \(\{\kappa(x_i)\}_{i=1}^m\).

As can be seen from \eqref{eq:deeponet-conventional}, the traditional DeepONet framework has certain limitations due to the structure of its branch network:
\begin{enumerate}
    \item DeepONet cannot handle situations where the sensor locations of the input function vary within the training or testing data. This inability to adapt to changing sensor positions limits its applicability.
    \item It is crucial that the sensor locations \(\{x_i\}_{i=1}^m\) in \(\myO\) are sufficiently numerous and effectively cover the domain of \(\kappa\) to enable accurate operator approximation (further theoretical analysis can be found in \cite{lanthaler2022error}). However, in many practical applications, the observations of the input function may be incomplete. This incompleteness, which can arise from sparse or noisy sensor data, may lead to potential inaccuracies in the operator approximation.
\end{enumerate}

VIDON, proposed by Prasthofer et al.~\cite{prasthofer2022variable}, address the first limitation by incorporating set transformer techniques~\cite{lee2019set}. However, VIDON assumes that the observational data is sufficiently informative to allow for a deterministic mapping from inputs to outputs. In this work, we extend this framework to handle incomplete observations and further account for the possibility that the operator itself exhibits inherent randomness—i.e., \(u\) may remain uncertain even if \(\kappa\) is fully specified.

To model the resultant uncertainties in prediction, we propose a stochastic operator model of the form:
\begin{equation}\label{eq:stochastic-operator}
\hat{\mathcal G}\left(h(\myO), z\right)(y) = \sum_{n=1}^{p} b^{n}\left(h(\myO), z\right) t^{n}(y),
\end{equation}
where \(h(\myO)\) is a set transformer~\cite{lee2019set} that provides a permutation-invariant representation of the observations \(\myO\), accommodating variable sensor counts and locations. The variable \(z\) is a latent random variable with a simple prior (e.g., standard Gaussian), capturing uncertainties arising from incomplete observations and inherent randomness in the operator.

In the following, we discuss how to train this model using data sampled from the joint distribution of input and output functions, aiming to accurately estimate the conditional distribution of \(u\) given \(\myO\).

\section{Methodology}
In this section, we develop UQ-SONet, a sensor permutation-invariant operator learning framework with UQ. The model consists of three main components: set transformer embedding, decoder, and encoder. The set transformer embedding adopts a structure similar to the VIDON architecture proposed in~\cite{prasthofer2022variable}, and integrates information from observations \( \myO \) of the input function \( \kappa \), including sensor coordinates and corresponding values, into a fixed-dimensional vector representation. This design enables the model to handle variable numbers and locations of sensors, while ensuring permutation invariance with respect to sensor ordering. The decoder generates the neural operator conditioned on the embedded observations and a random latent variable \( z \). The encoder estimates the approximate posterior of \( z \), conditioned on the observations and the known operator output \(u\), thereby providing an ELBO for the likelihood, which is used to train all model parameters.
Each component is described in detail below.

\subsection{Set transformer embedding}
The set transformer embedding consists of an input embedding module and an attention pooling module. Given a set of observations \(\myO = \{(x_i, \kappa(x_i))\}_{i=1}^m\) on \(\kappa\), the two modules are defined as follows:
\begin{itemize}
\item \textbf{Input embedding module}:
In this module, each sensor's position and corresponding value are independently mapped to vectors of equal dimensionality by two neural networks, $\Lambda_x$ and $\Lambda_{\kappa}$. The final combined embedding $\Lambda_i \in \mathbb{R}^{d_{emb}}$ for the $i$-th sensor is obtained by summing these two vectors:
\begin{equation}
    \begin{aligned}
            & \textbf{Combined embedding:} \quad \Lambda_i = \Lambda_x(x_i) + \Lambda_{\kappa}(\kappa(x_i)), \quad\text{for }i=1,\ldots,m.
        \end{aligned}
\end{equation}
Here, $d_{emb}$ denotes the dimensionality of the embedding.
      
\item \textbf{Attention pooling module}:  
This module consists of \(H\) attention heads. Each head comprises two trainable neural networks: an attention weights function \(w_l\) and a value projection function \(v_l\). These components enable the model to capture global dependencies across sensor observations and highlight the most informative locations. For any \(1 \leq l \leq H\), the computations are as follows:
\begin{equation}\label{eq:pooling}
	\begin{aligned}
		& \textbf{Single head output:} \quad\mathbb{R}^{d_{emb}} \rightarrow \mathbb{R}^{q}, \quad h^{(l)}(\myO) = \sum_{i=1}^{m} \underbrace{ \frac{\exp(w_l(\Lambda_i)/\sqrt{d_{emb}})}{\sum_{k=1}^{m} \exp(w_l(\Lambda_k)/\sqrt{d_{emb}})}}_{\text{attention weights}}  \underbrace{v_l(\Lambda_i)}_{\substack{\text{values} \\ \text{projection}}}.
	\end{aligned}
\end{equation}
By concatenating the tensors from each of the $H$ Head layers, we obtain the output $h\left(\myO\right)$, i.e.,
\begin{equation}\label{eq:set}
		h\left(\myO\right) = \left[h^{(1)}\left(\myO\right), h^{(2)}\left(\myO\right), ..., h^{(H)}\left(\myO\right)\right]^{\top}.
\end{equation}
\end{itemize}

Notice that the pooling operation defined in \eqref{eq:pooling} can process embeddings with varying numbers of sensor observations, and its output is invariant to their ordering. Consequently, the proposed set transformer embedding provides an effective way to encode the observation data $\myO$, representing the variable-sized set $\myO$ with a fixed-dimensional vector while strictly preserving permutation invariance.

\begin{figure}[!ht]
	\centering
	{\includegraphics[height=0.42\textheight, width=1.0\textwidth]{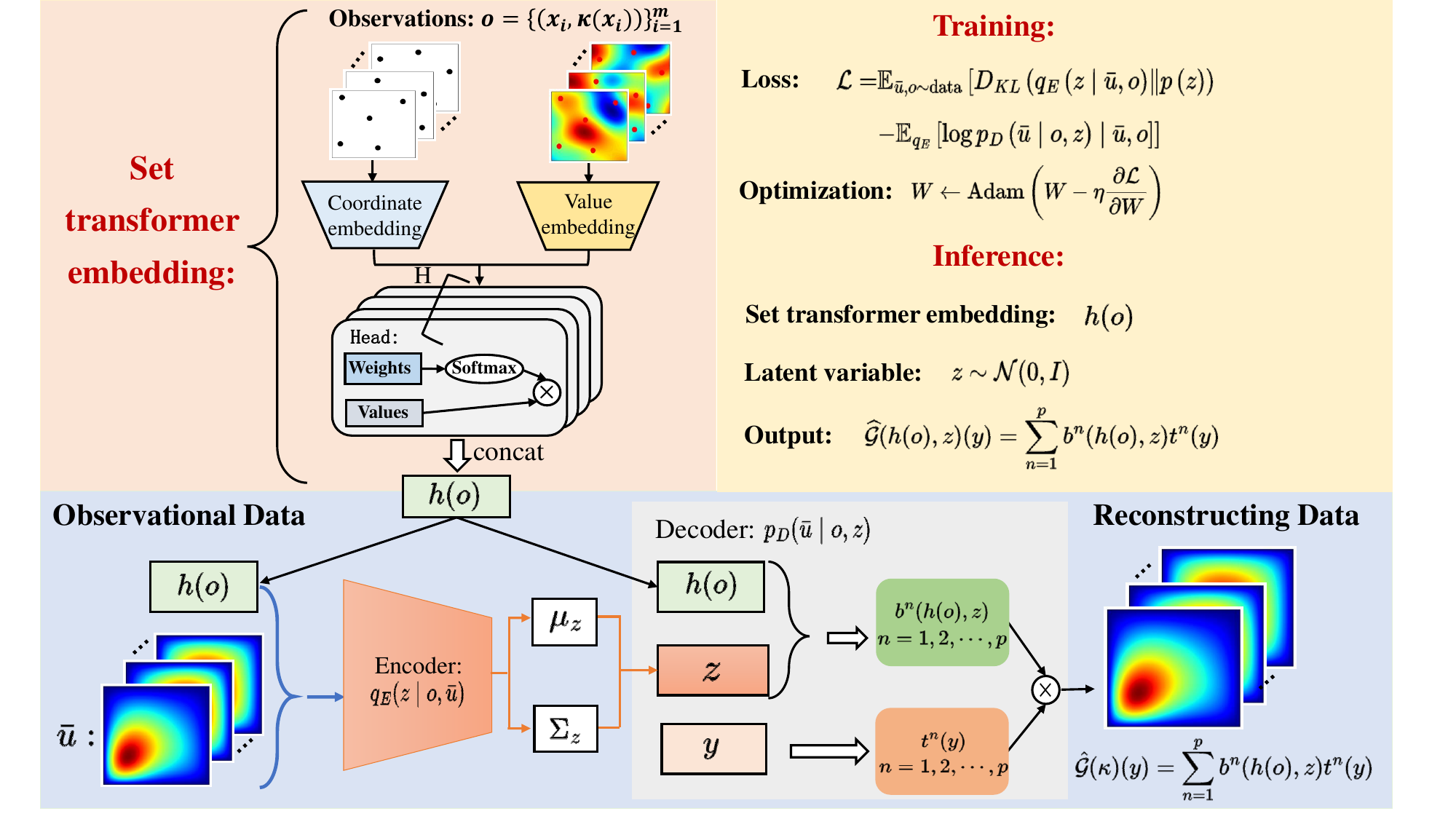}}\\
	\caption{\textbf{Schematic of the UQ-SONet model.} The top figure illustrates the set transformer embedding structure, where $h(\myO)$ is obtained by incorporating the coordinates and values of the input function sensors. The bottom figure depicts the encoder–decoder architecture, which derives the latent variables and reconstructs the target function of the operator.}\label{fig:setonet}
\end{figure}

\subsection{Encoder and decoder}

Functional VAE-type models remain one of the mainstream approaches in functional generative modeling. They are supported by a well-established theoretical foundation and a stable variational optimization framework, and continue to be actively studied in the recent literature \cite{bunker2025autoencoders, seidman2023variational}. Accordingly, we adopt the cVAE framework to approximate the conditional distribution under sparse and variable input observations. For computational convenience, we represent the function \( u \) by its values at a set of grid points \(\{y_1, \dots, y_M\}\), where \(\bar{u} = (u(y_1), \dots, u(y_M))^\top\). We assume that \( M \) is sufficiently large for \(\bar{u}\) to provide an adequate representation of \(u\). The decoder and encoder used for training are described as follows:\\
\begin{itemize}
  \item \textbf{Decoder}: Given the observations \( \myO \) and latent variable \( z \), the decoder defines the conditional distribution of $\bar u$ as
\begin{eqnarray}
p_{D}\left(\bar{u}\mid \myO,z\right) & = & \prod_{i=1}^{M}\mathcal{N}\left(u(y_{i})\mid\hat{\mathcal{G}}(h(\myO),z)(y_{i}),M\sigma_{u}^{2}\right)\nonumber\\
 & = & \prod_{i=1}^{M}\mathcal{N}\left(u(y_{i})\mid\sum_{n=1}^{p}b^{n}\left(h(\myO),z\right)t^{n}(y_{i}),M\sigma_{u}^{2}\right),\label{eq:decoder}
\end{eqnarray}
where \( p_D \) denotes the conditional likelihood defined by the decoder, and \(\mathcal{N}(\cdot \mid \mu, \Sigma)\) represents a Gaussian distribution with mean \(\mu\) and covariance \(\Sigma\). This formulation is based on the stochastic operator model \( \hat{\mathcal{G}} \) introduced in~\eqref{eq:stochastic-operator}, with an additional artificial noise term of variance \(\sigma_u^2\), which is a small positive hyperparameter. The prior distribution of \( z \) is set to the standard Gaussian, \( p(z) = \mathcal{N}(z \mid 0, I) \).
  \item \textbf{Encoder}: The encoder estimates the approximate posterior distribution of the latent variable \( z \) based on the observations \( \myO \) and the corresponding output \( \bar{u} \) from the decoder:
\[
q_E(z \mid \myO, \bar u) = \mathcal{N} \left( z \mid \mu_z(h(\myO), \bar{u}), \Sigma_z(h(\myO), \bar{u}) \right),
\]
where both \( \mu_z \) and \( \Sigma_z \) are modeled by neural networks, with \( \Sigma_z \) assumed to be diagonal, and $q_E$ denotes the conditional distribution defined by the encoder.
\end{itemize}

\begin{remark}
In the decoder~\eqref{eq:decoder}, the variance of the artificial noise is scaled with \( M \). This scaling ensures that as \( M \to \infty \), the resulting loss function can be interpreted as the loss of a cVAE defined in function space. See \eqref{eq:loss} and \ref{appendix:messure} for details.
\end{remark}

\subsection{Loss function}
In this section, we derive the loss function used to train the model. For any decoder, the conditional marginal likelihood of \( \bar{u} \) given observations \( \myO \) satisfies, by Jensen's inequality:
\begin{eqnarray*}
\log p_D(\bar{u} \mid \myO) 
& = & \log \int p_D(\bar{u} \mid \myO, z) p(z) \, \mathrm{d}z \\
& = & \log \int q_E(z \mid \bar{u}, \myO) \cdot \frac{p_D(\bar{u} \mid \myO, z) p(z)}{q_E(z \mid \bar{u}, \myO)} \, \mathrm{d}z \\
& \ge & \int q_E(z \mid \bar{u}, \myO) \cdot \log \frac{p_D(\bar{u} \mid \myO, z) p(z)}{q_E(z \mid \bar{u}, \myO)} \, \mathrm{d}z \\
& = & -D_{\mathrm{KL}}\left(q_E(z \mid \bar{u}, \myO) \,\|\, p(z)\right) + \mathbb{E}_{q_E} \left[ \log p_D(\bar{u} \mid \myO, z) \mid \bar{u}, \myO \right].
\end{eqnarray*}
The right-hand side provides a lower bound for \( \log p_D(\bar{u} \mid \myO) \), known as the evidence lower bound. Equality is achieved if and only if the approximate posterior \( q_E(z \mid \bar{u}, \myO) \) equals the true posterior \( p_D(z \mid \bar{u}, \myO) \).

Accordingly, we optimize the model via maximum likelihood estimation by minimizing the following loss function over the training dataset:
\begin{eqnarray}
	\mathcal{L} 
	& = & \mathbb{E}_{\bar{u}, \scalebox{0.6}{$\mathcal{O}$} \sim \mathrm{data}} \left[ D_{\mathrm{KL}}\left(q_E(z \mid \bar{u}, \myO) \,\|\, p(z)\right) - \mathbb{E}_{q_E} \left[\log p_D(\bar{u} \mid \myO, z) \mid \bar{u}, \myO \right] \right] \nonumber\\
	& = & \frac{1}{2} \, \mathbb{E}_{\bar{u}, \scalebox{0.6}{$\mathcal{O}$} \sim \mathrm{data}} \left[ -\log |\Sigma_{z}(h(\myO), \bar{u})| + \mathrm{tr}(\Sigma_{z}(h(\myO), \bar{u})) + \| \mu_{z}(h(\myO), \bar{u}) \|^2 - d_z \right] \nonumber\\
	& & + \frac{1}{2\sigma_u^2}\mathbb{E}_{\bar{u}, \scalebox{0.6}{$\mathcal{O}$} \sim \mathrm{data}, \, z \sim q_E(z \mid \bar{u}, \scalebox{0.6}{$\mathcal{O}$})} \left[ \sum_{i=1}^{M} \frac{ \left( \hat{\mathcal{G}}(h(\myO), z)(y_i) - u(y_i) \right)^2 }{M} \right],\label{eq:loss}
\end{eqnarray}
where $d_z$ denotes the dimension of the latent variable $z$, and the second equality follows from the fact that both \( p_D(\bar{u} \mid \myO, z) \) and \( q_E(z \mid \bar{u}, \myO) \) are Gaussian distributions. The loss \( \mathcal{L} \) provides an upper bound for the expected negative log-likelihood \( \mathbb{E}_{\bar{u}, \scalebox{0.6}{$\mathcal{O}$} \sim \mathrm{data}} [-\log p_D(\bar{u} \mid \myO)] \).

It is worth noting that when the grid points \( \{ y_1, \ldots, y_M \} \) are uniformly distributed and \( M \) is sufficiently large, the second term on the r.h.s.~of \eqref{eq:loss} becomes proportional to the mean squared \( L^2 \)-norm of the difference between the neural operator output \( \hat{\mathcal{G}}(h(\myO), z) \) and \( u \). In this case, \eqref{eq:loss} can be further interpreted as an approximation of the loss function for training a functional cVAE (see \ref{appendix:messure}).

A sketch of the computation graph for the UQ-SONet model is shown in Fig.~\ref{fig:setonet}. The training algorithm, as well as the inference procedure for function prediction and UQ, is summarized in Algorithm~\ref{alg:setonet}.

\begin{algorithm}[H]
	\caption{UQ-SONet for operator learning.}
	\label{alg:setonet}
	\textbf{Training stage:}\\
	\textbf{Input:} Training set \( \{(\kappa_i, u_i)\}_{i=1}^N \);  
Range of allowable sensor counts \( \mathcal{M} \subset \mathbb{Z}^+ \);  
Generated measurements of the output function \( u \) at fixed grid points, \( \{ \bar{u}_i \}_{i=1}^N = \left\{ \left\{ u_i(y_j) \right\}_{j=1}^M \right\}_{i=1}^N \); batch size $B$;
	
	\begin{enumerate}
		\item Initialize all neural networks, including:  
\( \Lambda_x \), \( \Lambda_{\kappa} \), \( \{ w_l \}_{l=1}^H \), \( \{ v_l \}_{l=1}^H \) in the set transformer embedding;  
\( \{ b^n \}_{n=1}^p \), \( \{ t^n \}_{n=1}^p \) in the decoder;  
\( \mu_z \) and \( \Sigma_z \) in the encoder;

        \item Randomly sample a mini-batch \( \{ (\kappa_i, \bar{u}_i) \}_{i=1}^B \) from the training dataset;

        \item Randomly select \( m \in \mathcal{M} \), and generate a set of observations \( \myO_i = \{ (x_j^i, \kappa_i(x_j^i)) \}_{j=1}^m \) for \( i = 1, \ldots, B \), where each \( x_j^i \) is randomly sampled from the domain of \( \kappa \);

		\item Calculate $h(\myO_1),\ldots,h(\myO_B)$ through the set transformer embedding \eqref{eq:set};
		
		\item Draw $z_i\sim q_E(z_i|\myO_i,\bar u_i)$ for $i=1,\ldots,B$;

		\item Compute a stochastic approximation of the loss function:
        \begin{eqnarray*}
\hat{\mathcal{L}} & = & \frac{1}{2B}\sum_{i=1}^{B}\left(-\log|\Sigma_{z}(h(\myO_{i}),\bar{u}_{i})|+\mathrm{tr}(\Sigma_{z}(h(\myO_{i}),\bar{u}_{i}))+\|\mu_{z}(h(\myO_{i}),\bar{u}_{i})\|^{2}-d_{z}\right)\\
 &  & +\frac{1}{2BM\sigma_{u}^{2}}\sum_{i=1}^{B}\sum_{j=1}^{M}\left(\hat{\mathcal{G}}(h(\myO_{i}),z_{i})(y_{j})-u(y_{j})\right)^{2};
\end{eqnarray*}
		
		\item Update all trainable parameters $W$ of neural networks as: 
		\[
		W \leftarrow \text{Adam}\left(W - \eta\frac{\partial\hat{\mathcal{L}}}{\partial W}\right),
		\]
		where $\eta$ is the learning rate;

        \item Repeat Steps 2-7 until convergence.
	\end{enumerate}
	\textbf{Inference:}\\
	\textbf{Input}: $m$ sensors and corresponding measurements at arbitrary positions of the input function: $\myO = \{x_i, \kappa(x_i)\}^{m}_{i=1}$, $m \in \mathcal{M}$;
	\begin{enumerate}
		\item Compute $h(\myO)$ by \eqref{eq:set};
		
		\item Sample the latent variable $z\sim\mathcal{N}(0,I)$;
		
		\item Calculate the prediction $\hat{\mathcal G}\left(h(\myO), z\right)$ of the output function by \eqref{eq:stochastic-operator};
		
		\item Repeat Steps 2-3 more times to calculate the quantities of interests (QoI) of the conditional distribution of $u$.
	\end{enumerate}
\end{algorithm}

\section{Simulation Results}\label{sec:results}

In this section, we evaluate the performance of the proposed UQ-SONet framework through a series of benchmark problems. To illustrate its flexibility, we adopt the variable random sensor configurations introduced in \cite{prasthofer2022variable}, where both the number and positions of input sensors are allowed to vary arbitrarily. In addition, we evaluate scenarios with noisy input observations to further demonstrate the robustness of our method. We consider three representative classes of operators: deterministic operators, stochastic operators, and a time-dependent operator arising from the Navier–Stokes equations. For the benchmark problems in Sections \ref{subsec:1d-pde}–\ref{subsec:2d-sde}, we use the hard constraint construction from \cite{lu2021physics}:
\begin{equation*}
    u(x)=(1-x)(1+x)\,u_{\theta}(x),
\end{equation*}
in one dimension, and
\begin{equation*}
    u(x,y)=x(1-x)\,y(1-y)\,u_{\theta}(x,y),
\end{equation*}
in two dimensions. Here $u_{\theta}$ denotes the unconstrained neural output (i.e., the model prediction before boundary enforcement) for each example. The multiplicative factors enforce homogeneous Dirichlet conditions directly by construction.

For all experiments, we employ the hyperbolic tangent (Tanh) activation function in the neural networks, and set the truncation parameter of the operator network to
$p=100$. The hidden-layer sizes and other model hyperparameters for each case are summarized in Table \ref{tab:NN}. The computational costs of all experiments, including the training and inference times, are summarized in Table \ref{tab:training-cost} in \ref{appendix:cost}. Model accuracy is evaluated using two complementary metrics: the Wasserstein-2 distance \cite{mallasto2017learning}, defined as
\begin{equation}
	W_2(\mu, \nu) = \left( \inf_{\gamma \in \Gamma(\mu, \nu)} \int_{M \times M} d^2(x_1, x_2) d\gamma(x_1, x_2) \right)^{1/2}, \quad (x_1, x_2) \in M \times M,
\end{equation}
where $(M, d)$ is a Polish space (i.e., a complete and separable metric space), and $\Gamma(\mu, \nu)$ denotes the set of joint measures between the probability measures $\mu$ and $\nu$ on $M$; relative errors of the mean and standard deviation for the estimated conditional distribution, defined as:
\begin{equation}
	\begin{aligned}
		\text { Relative error in mean } & =\frac{\|E[u]-E_{\theta}[u]\|_2}{\|E[u]\|_2}, \\
		\text { Relative error in std } & =\frac{\|\sigma[u]-\sigma_{\theta}[u]\|_2}{\|\sigma[u]\|_2},
	\end{aligned}
\end{equation}
where $E[u]$ and $\sigma[u]$ denote the mean and standard deviation of the reference solutions computed by the method presented in \ref{appendix:reference}, respectively, and $E_{\theta}[u]$ and $\sigma_{\theta}[u]$ denote the corresponding predictions obtained from UQ-SONet. To assess the robustness of the model, all experiments are repeated with multiple independent trials. We report the averaged errors of the mean and standard deviation, denoted by $err_{E[u]}$ and $err_{\sigma[u]}$, respectively. 

\subsection{One-dimensional Diffusion Equation.}\label{subsec:1d-pde}
We first study the one-dimensional diffusion equation:
\begin{equation}
	\begin{aligned}\label{eq:1d}
		-\frac{1}{10}\frac{d}{dx}(k(x) \frac{du}{dx} ) & = f(x), \quad x \in \left[-1, 1\right], \\
		u\left(-1\right) = u\left(1\right) & = 0, 
	\end{aligned}
\end{equation}
where $k$ is the diffusion coefficient and $f(x) = 2\sin(2\pi x)$ is the source term. The goal is to learn the operator $\mathcal{G}: k(x) \longmapsto u(x)$. 

The training and testing data are generated as follows. To ensure positivity of the coefficient, the log function of the input function $k$ is sampled from the following Gaussian process
\begin{equation}\label{eqn:1d-gp}
	\begin{aligned}
		\log{k}\sim\mathcal{G P}\left(\mu(x), \sigma^2 \exp \left(-\frac{\left(x-x^{\prime}\right)^2}{l^2}  \right)\right), \\
	\end{aligned}
\end{equation}
with $\mu(x) = \sin(2\pi x)$, fluctuation amplitude $\sigma=0.5$, and correlation length $l=0.1$. Each realization of $k$ is recorded on an equally spaced grid with $401$ sensor locations. The diffusion equation is solved using a second-order finite difference scheme to obtain the corresponding solution $u(x)$. In total, we generate $N = 10,000$ input/output function pairs $ \{(k_i, u_i)\}_{i=1}^N$. Although each solution $u$ is computed on $401$ grid points, only 101 uniformly spaced points are used during training.

We first consider the case where the number of observations is fixed while their locations vary. In this setting, we investigate the effects of the latent dimension, the size of the training dataset, and the number of input sensors on the performance of our method. The reference solution is computed using the conditional Gaussian process method, with details provided in \ref{appendix:reference}. Fig. \ref{fig:1d-pde-compare}(a) reports the Wasserstein-2 distance between the reference solution and the conditional distribution learned by UQ-SONet under different latent dimensions. We observe that the Wasserstein-2  distance stabilizes once the latent dimension reaches 10; therefore, we use a latent dimension of 10 in all subsequent experiments. Fig. \ref{fig:1d-pde-compare}(b) presents the relative $L_2$ errors of the predicted mean and standard deviation across different training dataset sizes, where sensor configurations of 3 and 5 are evaluated separately. The results show that larger training datasets yield smaller errors. However, considering both data generation and computational cost, we use 10,000 samples in all following examples. Fig. \ref{fig:1d-pde-compare}(c) illustrates the relationship between the mean of the standard deviations of the predicted solutions and the number of input sensors, under a fixed latent dimension of 10 and a training set size of 10,000 samples. These results demonstrate that UQ-SONet can effectively quantify the uncertainty of the output function and that the output uncertainty decreases as more observational information from the input function becomes available. 

\begin{figure}[!ht]
	\centering
	{\includegraphics[height=0.40\textheight, width=1.0\textwidth]{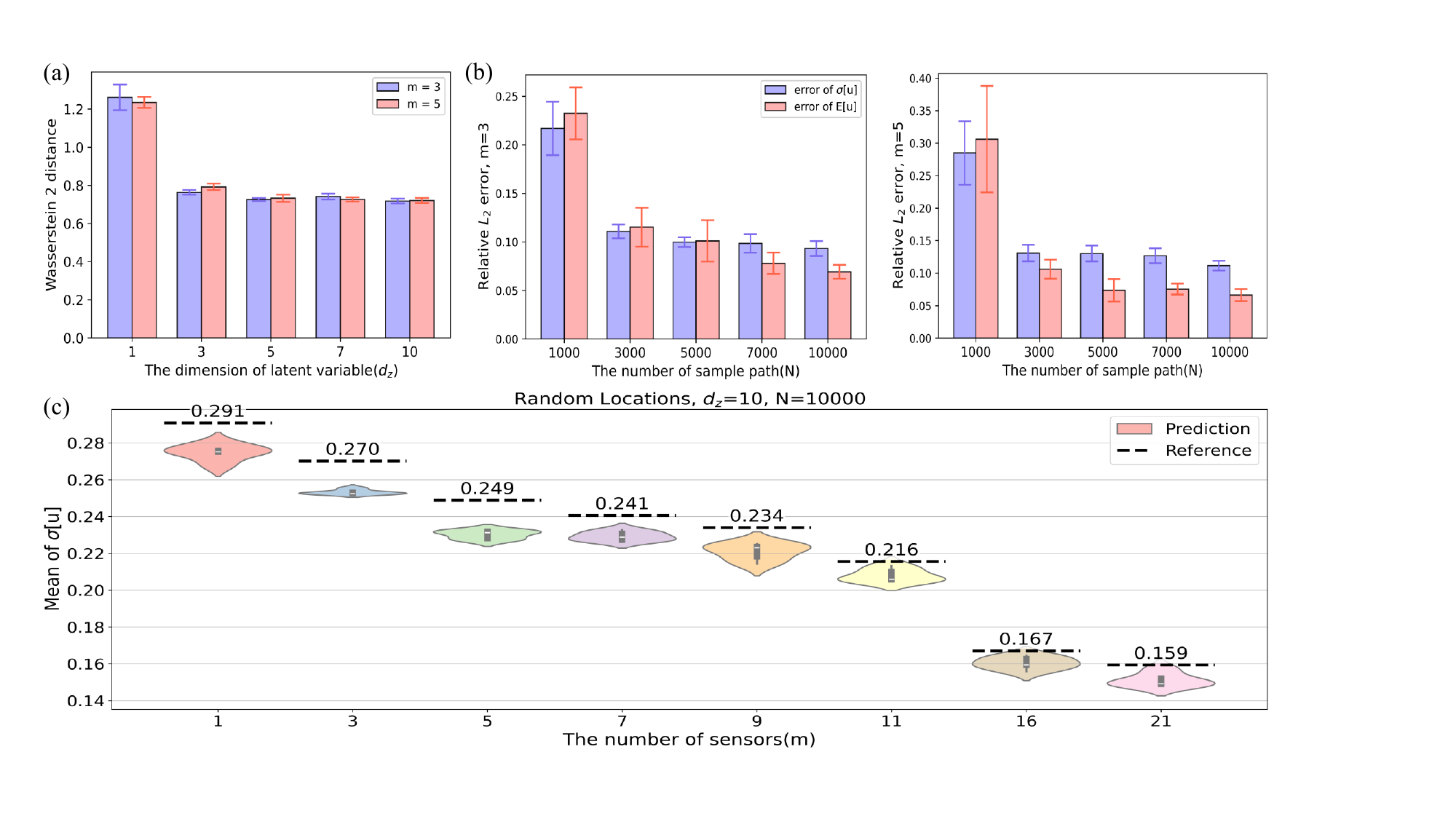}}\\
	\caption{\textbf{Diffusion Equation \eqref{eq:1d}.} \textbf{(a):} Wasserstein-2 (W2) distance between the reference distribution and the conditional distribution obtained by UQ-SONet, with $m=3$ and $m=5$ (red). \textbf{(b):} Relative $L_2$ errors of the mean and standard deviation between UQ-SONet predictions and reference values with $m=3$ (left) and $m=5$ (right); \textbf{(c):} Mean of the standard deviations of conditional distributions generated by UQ-SONet compared with reference values. Bars indicate results averaged over five independent experiments.}\label{fig:1d-pde-compare}
\end{figure}

\begin{figure}[!ht]
	\centering
	{\includegraphics[height=0.40\textheight, width=1.0\textwidth]{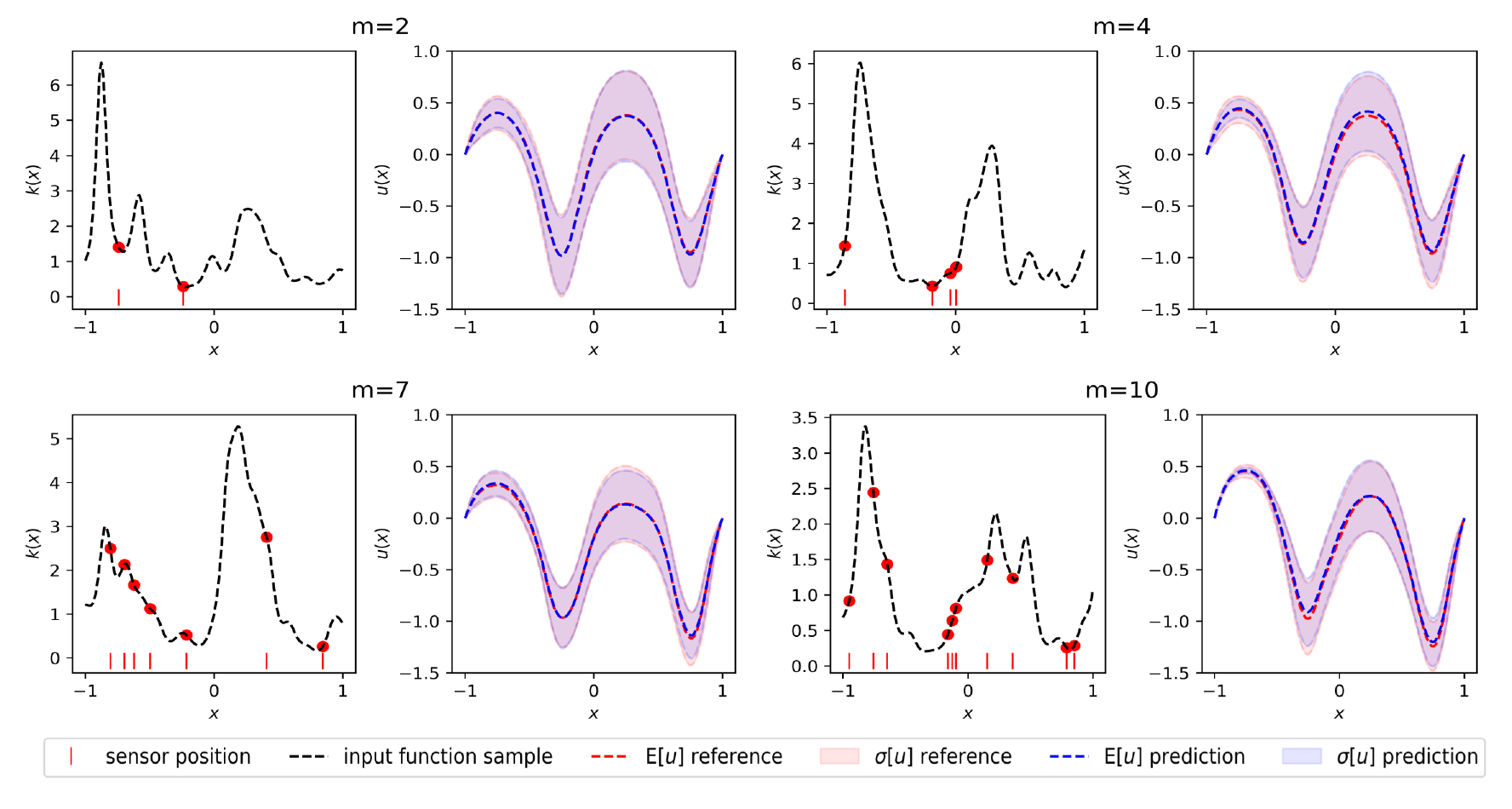}}\\
	\caption{\textbf{Diffusion Equation \eqref{eq:1d}.} Reference solution versus the prediction of UQ-SONet for a representative example with $m=2$ (top left), $m=4$ (top right), $m=7$ (bottom left), $m=10$ (bottom right). Black dashed lines show sampled input values, and red markers indicate sensor locations. Red and blue dashed lines with shaded regions denote the mean and standard deviation of the reference and predicted solutions, respectively. }\label{fig:1d-pde-plot}
\end{figure}

We further test the case where both the number and the locations of sensors are allowed to vary. The set of sensor counts for the input function is defined as $\mathcal{M} = \{1, 2, 3, 4, 5, 6, 7, 8, 9, 10\}$. During training, the total data set is evenly divided into $10$ batches, each containing $1,000$ samples. For each batch, we randomly select a number $m\in \mathcal{M}$ and use it to generate input observations by sampling sensor locations $x$ at different positions together with their corresponding values of $k(x)$. The associated output measurements are uniformly sampled at $101$ grid points. To reduce randomness, we pregenerate $10$ batches and randomly select one batch at each iteration for model optimization. All models are trained for $100,000$ iterations. For testing, we construct a dataset of $10$ pairs of input-output functions. Input observations are generated in the same manner as in training, with sensor counts selected from $\mathcal{M}$. The corresponding output functions are evaluated on $401$ uniformly spaced grid points across the domain. We report both the mean predictions and the standard deviations for representative test cases with different number of sensors in Fig. \ref{fig:1d-pde-plot}. For illustration, the number of input function sensors is set to $2$, $4$, $7$, and $10$, respectively. As shown in the figure, the mean and standard deviation produced by our UQ-SONet align closely with the reference solutions, demonstrating the accuracy and reliability of the proposed framework. Moreover, as the number of sensors increases, the predictive uncertainty bands gradually shrink, indicating that additional observational information leads to more confident and accurate predictions. For completeness, we also include a comparison with a deep ensemble baseline based on VIDON, with detailed results and discussion deferred to \ref{appendix:deepensem}.

 Next, we compare our method with the VIDON framework. To ensure fair training for both models and, in particular, to prevent overfitting in VIDON, which requires a sufficiently large number of sensors in its original formulation, we adopt the following strategy. At each training iteration, we first randomly select a sensor count and then randomly choose a batch of $1,000$ samples, where both the number and the positions of sensors are kept consistent within that batch. For each case, we conduct five independent training trials and report the mean and standard deviation of the errors to demonstrate both the accuracy and robustness of our model in Table \ref{tab:1d-pde-error}. The results show that under insufficient sensor inputs, our UQ-SONet not only achieves higher precision in mean estimation compared to the deterministic VIDON method, but also consistently produces stable conditional distributions for individual samples of the output function. This highlights the advantage of incorporating UQ into operator learning, particularly under sparse and variable sensor observations.

\begin{table}[!ht]
	\centering
	\small
	\setlength{\tabcolsep}{9pt} 
	\renewcommand{\arraystretch}{1.45} 
	\caption{\textbf{ Diffusion Equation \eqref{eq:1d}.} Relative $L_2$ errors of the mean and standard deviation between the reference distribution and the conditional distribution predicted by UQ-SONet (with comparison to VIDON, which provides only mean estimates). The $\pm$ values are computed from five independent experiments. All reported values are scaled by a factor of $1 \times 10^{-2}$.}\label{tab:1d-pde-error}
	\begin{tabular}{ c | c c c | c | c c c }
		\hline
		\multirow{2}{*}{\centering \makecell{$m$}} & \textbf{VIDON} & \multicolumn{2}{c|}{\textbf{UQ-SONet}} & \multirow{2}{*}{\centering \makecell{$m$}} & \textbf{VIDON} & \multicolumn{2}{c}{\textbf{UQ-SONet}} \\ 
		\cline{2-4} \cline{6-8}
		& $err_{E[u]}$ & $err_{E[u]}$ & $err_{\sigma[u]}$ & & $err_{E[u]}$ & $err_{E[u]}$ & $err_{\sigma[u]}$ \\ 
		\hline
		1  & $5.35 \pm 0.15$ & $4.47 \pm 0.83$ & $8.77 \pm 0.86$ & 6  & $7.14 \pm 0.40$ & $5.67 \pm 0.68$ & $8.74 \pm 0.58$ \\ 
		2  & $5.64 \pm 0.99$ & $4.62 \pm 0.54$ & $8.73 \pm 0.49$ & 7  & $8.73 \pm 1.03$ & $5.88 \pm 0.74$ & $8.95 \pm 0.52$ \\ 
		3  & $6.17 \pm 0.91$ & $4.81 \pm 0.37$ & $8.57 \pm 0.53$ & 8  & $6.36 \pm 0.59$ & $5.93 \pm 0.44$ & $7.68 \pm 0.55$ \\ 
		4  & $6.98 \pm 0.80$ & $5.52 \pm 0.80$ & $7.79 \pm 0.85$ & 9  & $7.92 \pm 0.61$ & $5.99 \pm 0.74$ & $8.09 \pm 0.44$ \\ 
		5  & $5.41 \pm 0.79$ & $5.40 \pm 0.26$ & $8.48 \pm 0.45$ & 10 & $7.10 \pm 1.16$ & $5.97 \pm 0.37$ & $8.35 \pm 0.38$ \\ 
		\hline
	\end{tabular}
\end{table}

\begin{figure}[!ht]
	\centering
	{\includegraphics[height=0.40\textheight, width=1.0\textwidth]{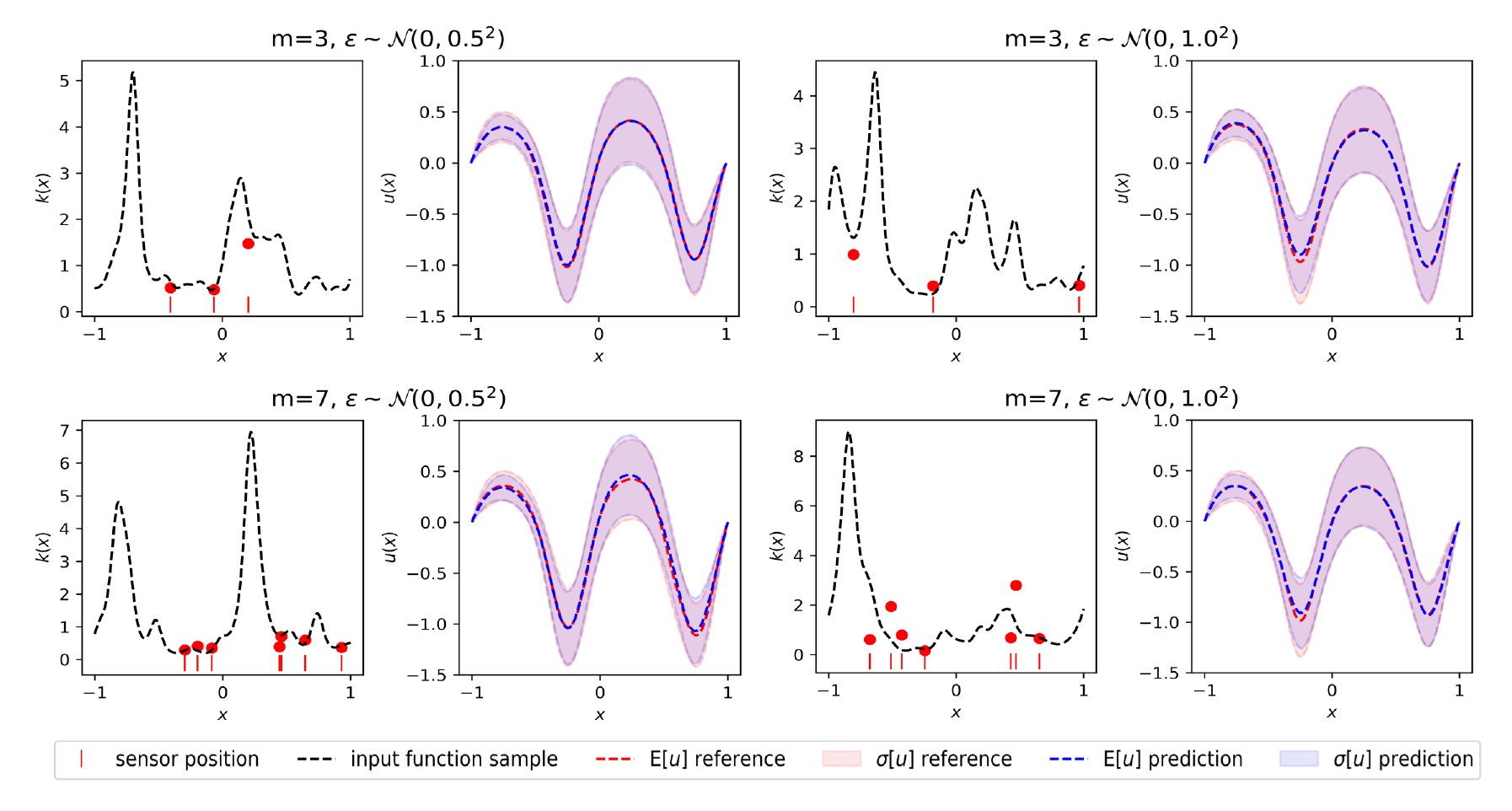}}\\
    \caption{\textbf{Diffusion Equation \eqref{eq:1d}.} Mean and standard deviation of the conditional distribution obtained by UQ-SONet compared with reference values for $3$ (top) and $7$ (bottom) input sensors under noise conditions: $\epsilon \sim \mathcal{N}(0, 0.5^2)$ (left) and $\epsilon \sim \mathcal{N}(0, 1.0^2)$ (right). Red dots mark the noise-contaminated observations at the corresponding sensor locations. Red and blue curves denote the reference and predicted solutions, respectively, while black dashed lines indicate input samples. }\label{fig:1d-pde-addnoise}
\end{figure}

To comprehensively evaluate the robustness and generalization ability of the proposed method, we further examine the scenario where the observations of the input functions $k$ are corrupted by multiplicative noise. We consider a fixed-sensor setting with variable locations, using 3 and 7 sensors as representative cases. During both training and testing, the observation provided by a sensor at location $x$ is $k(x)\cdot\exp(\epsilon(x))$, where $\epsilon(x)\sim\mathcal{N}(0, \sigma^2)$ denotes spatially independent and identically distributed noise. We investigate five noise levels with $\sigma \in \{0.1, 0.3, 0.5, 0.7, 1.0\}$. The results are presented in Fig. \ref{fig:1d-pde-addnoise}, and the relative $L_2$ errors of the mean and standard deviation are summarized in Table \ref{tab:1d-pde-addnoise-error}.

\begin{table}[!ht]
	\centering
	\small
	\setlength{\tabcolsep}{12pt} 
	\renewcommand{\arraystretch}{1.55} 
	\caption{\textbf{Diffusion Equation \eqref{eq:1d}.} Relative $L_2$ errors of the mean and standard deviation between the reference  distribution and the conditional distribution predicted by UQ-SONet under different input noise levels for 3 and 7 sensors.  The $\pm$ is calculated from $5$ independent experiments. All values are scaled by a factor of $1 \times 10^{-2}$.}
	\label{tab:1d-pde-addnoise-error}
	\begin{tabular}{ c | c c | c c c c c }
		\hline
		\multirow{2}{*}{\centering \makecell{$\sigma$}} & \multicolumn{2}{c}{\makecell{$\mathbf{m = 3}$}} & \multicolumn{2}{c}{\centering \makecell{$\mathbf{m = 7}$}}\\ 
		\cline{2-5}
		& $err_{E[u]}$ & $err_{\sigma[u]}$ & $err_{E[u]}$ & $err_{\sigma[u]}$\\ 
		\hline
		0.1  & $6.77 \pm 1.17$ & $9.43 \pm 0.48$ & $7.49 \pm 0.95$ & $9.67 \pm 0.32$ \\ 
		0.3  & $7.06 \pm 1.73$ & $9.46 \pm 0.46$ & $7.08 \pm 1.71$ & $9.31 \pm 0.75$ \\ 
		0.5  & $6.69 \pm 0.65$ & $9.07 \pm 0.47$ & $6.04 \pm 0.67$ & $9.02 \pm 0.42$ \\ 
		0.7  & $7.16 \pm 0.45$ & $8.90 \pm 0.71$ & $7.11 \pm 1.48$ & $9.35 \pm 0.80$ \\ 
		1.0  & $6.58 \pm 1.03$ & $9.20 \pm 1.07$ & $6.50 \pm 0.69$ & $7.52 \pm 0.35$ \\
		\hline
	\end{tabular}
\end{table}

\subsection{Two-dimensional Poisson Equation.}\label{subsec:2d-pde}
Next, we consider the following two-dimensional poisson equation defined on the domain $\mathcal{D} = [0,1]^2$ with Dirichlet boundary conditions:
\begin{equation}\label{eqn:2d-pde}
	\begin{aligned}
		-\frac{1}{10} \Delta u\left(x, y\right) & = f\left(x, y\right), \quad \left(x, y\right) \in \mathcal{D}, \\
		u\left(x, y\right) & = 0, \quad \left(x, y\right) \in \partial \mathcal{D}.
	\end{aligned}
\end{equation}
Here, the goal is to learn the operator that maps the source term $f$ to the solution $u$, that is, $\mathcal{G}: f(x, y) \longmapsto u(x, y)$. The source term $f$ is sampled from a two-dimensional Gaussian process \cite{lu2022comprehensive}:

\begin{equation}\label{eqn:2d-gp}
	\begin{aligned}
		f\sim\mathcal{G P}\left(\mu(x), \exp \left[\frac{-\left(x-x^{\prime}\right)}{2 l_1^2}+\frac{-\left(y-y^{\prime}\right)^2}{2 l_2^2}\right]\right), \\
	\end{aligned}
\end{equation}
with $\mu(x,y) = 4(sin(2 \pi x) + sin(2 \pi y))$, and the correlation lengths are set to $l_1=l_2=0.1$. Each realization of $f$ is evaluated on a uniformly spaced $101 \times 101$ grid. Then, the corresponding output functions $u$ are computed using a second-order finite difference scheme on a same grid, following the numerical implementation in \cite{li2020fourier}. We generate $N=80,000$ input/output function pairs $\{(f_i, u_i)\}_{i=1}^{N}$ to train the UQ-SONet model. To evaluate performance, additional input-output pairs $10$ are generated in the same way and are used as the test set.

\begin{figure}[!ht]
	\centering
	{\includegraphics[height=0.34\textheight, width=1.0\textwidth]{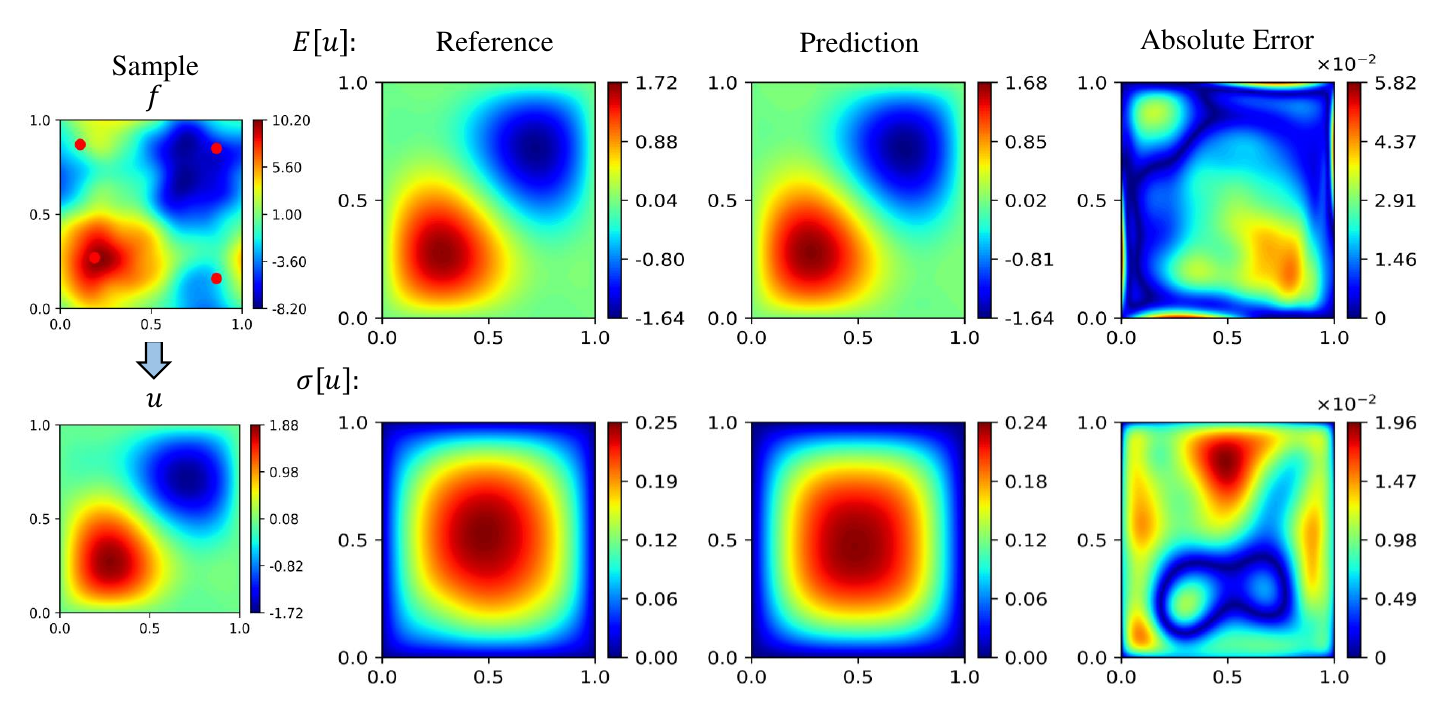}}\\
	\caption{\textbf{2d Poisson Equation \eqref{eqn:2d-pde}.} \textbf{Left:} reference samples of the two-dimensional operator map. Red markers denote the sensor locations. \textbf{Top row:} reference mean, UQ-SONet predicted mean, and the corresponding absolute error. \textbf{Bottom row:} reference standard deviation, UQ-SONet predicted standard deviation, and the corresponding absolute error.}\label{fig:2d-pde-plot}
\end{figure}

We next describe the training setup and evaluation metrics for this 2D problem. We consider both the number and the locations of sensors are allowed to vary and the set of sensor counts $\mathcal{M} = \{1, 2, 3, 4\}$. We pregenerate $4$ batches of observations, each containing $20,000$ samples, following a procedure similar to that described in Section \ref{subsec:1d-pde}. The only difference lies in the sampling of sensor locations, which are selected using a regular space clustering method \cite{prinz2011markov} applied to randomly permuted $81 \times 81$ uniform grids within the subdomain $[0.1, 0.9]^2$. This procedure ensures that the sensors uniformly and efficiently cover the domain while maintaining a minimum spacing of at least $d_{\min}$ between any two sensors. Specifically, we set $d_{\min}=0.8$ when the number of sensors is two, and $d_{\min}=0.5$ when the number of sensors is three or four. In our implementation, the regular space clustering is performed using the \lstinline|pyemma.coordinates.cluster_regspace| function from the PyEMMA package \cite{scherer2015pyemma}. For the output function $u$, we uniformly sample measurements on a grid $M=51 \times 51$. At each training iteration, a single batch is randomly selected for model optimization, and the model is trained for a total of $20,000$ iterations. During testing, the output function is predicted on a uniformly spaced $101 \times 101$ grid over the domain, using input observations generated in the same manner as during training.

We now evaluate the performance of UQ-SONet under both noise-free and noisy input observations. For the noisy case, Gaussian noise is added to the sensor measurements with levels $\mathcal{N}(0, 0.5^2)$ and $\mathcal{N}(0, 1.0^2)$. Fig. \ref{fig:2d-pde-plot} presents the conditional distribution results for a representative sample, where the input function is observed with four sensors and corrupted by Gaussian noise $\epsilon \sim \mathcal{N}(0, 1.0^2)$. Table \ref{tab:2d-pde-error} summarizes the relative $L_2$ errors of the predicted mean and standard deviation for both noise-free and noisy scenarios. Each result is averaged over three independent experiments to assess the robustness of the method. For the noiseless observation case, a comparison with the deterministic VIDON approach is also provided. The results demonstrate that UQ-SONet provides accurate and stable UQ for the output function across all scenarios. In particular, even when the number of input sensors is small and the observations are corrupted by substantial noise, the model remains capable of producing reliable conditional distributions, highlighting its robustness in handling noisy and limited data.

\begin{table}[!ht]
	\centering
	\small
	\setlength{\tabcolsep}{8pt} 
	\renewcommand{\arraystretch}{1.45} 
	\caption{\textbf{2d Poisson Equation \eqref{eqn:2d-pde}.} Relative $L_2$ errors of the mean and standard deviation between the reference and predicted distributions obtained by UQ-SONet (with comparison to VIDON, which provides only mean estimates for noiseless observations), reported for noiseless observations and for noisy observations with additive Gaussian noise $\mathcal{N}(0, 0.5^2)$ and $\mathcal{N}(0, 1.0^2)$. The $\pm$ values are computed from $3$ independent experiments. All reported values are scaled by a factor of $1 \times 10^{-2}$. }
	\label{tab:2d-pde-error}
	\begin{tabular}{ c | c | c c | c c | c c }
		\hline
		\multirow{3}{*}{\centering \makecell{$m$}} & \textbf{VIDON} & \multicolumn{6}{c}{\textbf{UQ-SONet}} \\ 
		\cline{2-8}
		& \makecell{noiseless} & \multicolumn{2}{c}{\makecell{noiseless}} & \multicolumn{2}{c}{\centering \makecell{$\epsilon \sim \mathcal{N}(0, 0.5^2)$}} & \multicolumn{2}{c}{\centering \makecell{$\epsilon \sim \mathcal{N}(0, 1.0^2)$}}\\ 
		\cline{2-8}
		& $err_{E[u]}$ & $err_{E[u]}$ & $err_{\sigma[u]}$ & $err_{E[u]}$ & $err_{\sigma[u]}$ & $err_{E[u]}$ & $err_{\sigma[u]}$ \\ 
		\hline
		1  & $3.26 \pm 0.19$ & $4.00 \pm 0.35$ & $9.46 \pm 0.28$ & $3.40 \pm 0.60$ & $9.22 \pm 0.60$ & $3.40 \pm 0.52$ & $8.66 \pm 0.37$\\ 
		2  & $4.67 \pm 0.24$ & $4.83 \pm 0.01$ & $8.75 \pm 0.43$ & $3.86 \pm 0.30$ & $8.30 \pm 0.81$ & $3.40 \pm 0.17$ & $8.79 \pm 0.26$\\ 
		3  & $5.99 \pm 0.88$ & $5.92 \pm 0.30$ & $9.18 \pm 0.05$ & $4.53 \pm 0.22$ & $8.61 \pm 0.22$ & $4.16 \pm 0.41$ & $7.49 \pm 0.03$\\ 
		4  & $7.01 \pm 1.15$ & $5.48 \pm 0.16$ & $9.37 \pm 0.01$ & $5.03 \pm 0.15$ & $8.01 \pm 0.26$ & $5.54 \pm 0.53$ & $7.27 \pm 0.21$\\ 
		\hline
	\end{tabular}
\end{table}

\subsection{One-dimensional  Stochastic Differential Equation.}\label{subsec:1d-sde}
To further demonstrate the advantage of our UQ-SONet model, we consider the following one-dimensional elliptic stochastic differential equation:
\begin{equation}
	\begin{aligned}\label{eq:1d-sto}
		-\frac{1}{10}\frac{d}{dx}(k(x; \omega) \frac{d}{dx}u(x; \omega) ) & = f(x;\omega), \quad x \in [-1, 1], \quad \omega \in \Omega, \\
		u(-1) = u(1) & = 0.
	\end{aligned}
\end{equation}
Here both the diffusion coefficient $k(x; \omega)$ and the source term $f(x; \omega)$ are modeled as independent stochastic processes. The logarithms of $k(x; \omega)$ and $f(x; \omega)$ are assumed to follow Gaussian processes defined by \eqref{eqn:1d-gp} with the following hyperparameters:
\begin{itemize}
	\item For $\text{log}~k(x; \omega)$: $l = 0.05$, $\sigma=0.3$ and $\mu(x) = sin(\pi x + 1)$;	
	\item For $f(x; \omega)$: $l = 0.1$, $\sigma = 0.1$, and $\mu(x) = \sin(2\pi x) + 0.1$.
\end{itemize}

In this example, we aim to learn the operator: $\mathcal{G}: k(x; \omega) \longmapsto u(x; \omega)$. Unlike the deterministic case studied in Section~\ref{subsec:1d-pde}, however, the uncertainty in the solution $u$ now arises not only from the insufficient observations of $k$ but also from the additional randomness introduced by the stochastic process $f$. This setting provides a more comprehensive test of the effectiveness of the proposed model.

For this stochastic elliptic problem, the training and testing datasets are generated following the same procedure described in Section~\ref{subsec:1d-pde}. Each realization of the stochastic processes $k(x; \omega)$ and $f(x; \omega)$ is discretized on a uniform grid, and the corresponding solution 
$u(x; \omega)$ is obtained using a second-order finite difference scheme. For the output function 
$u$, we uniformly sample measurements on a 101 point grid during training, while predictions are evaluated on a 401 point grid. During training, the data set is divided into batches of 1,000 samples, and at each iteration, a batch is randomly selected for model optimization. All models are trained for 100,000 iterations.

\begin{figure}[!ht]
	\centering	
	{\includegraphics[height=0.30\textheight, width=0.46\textwidth]{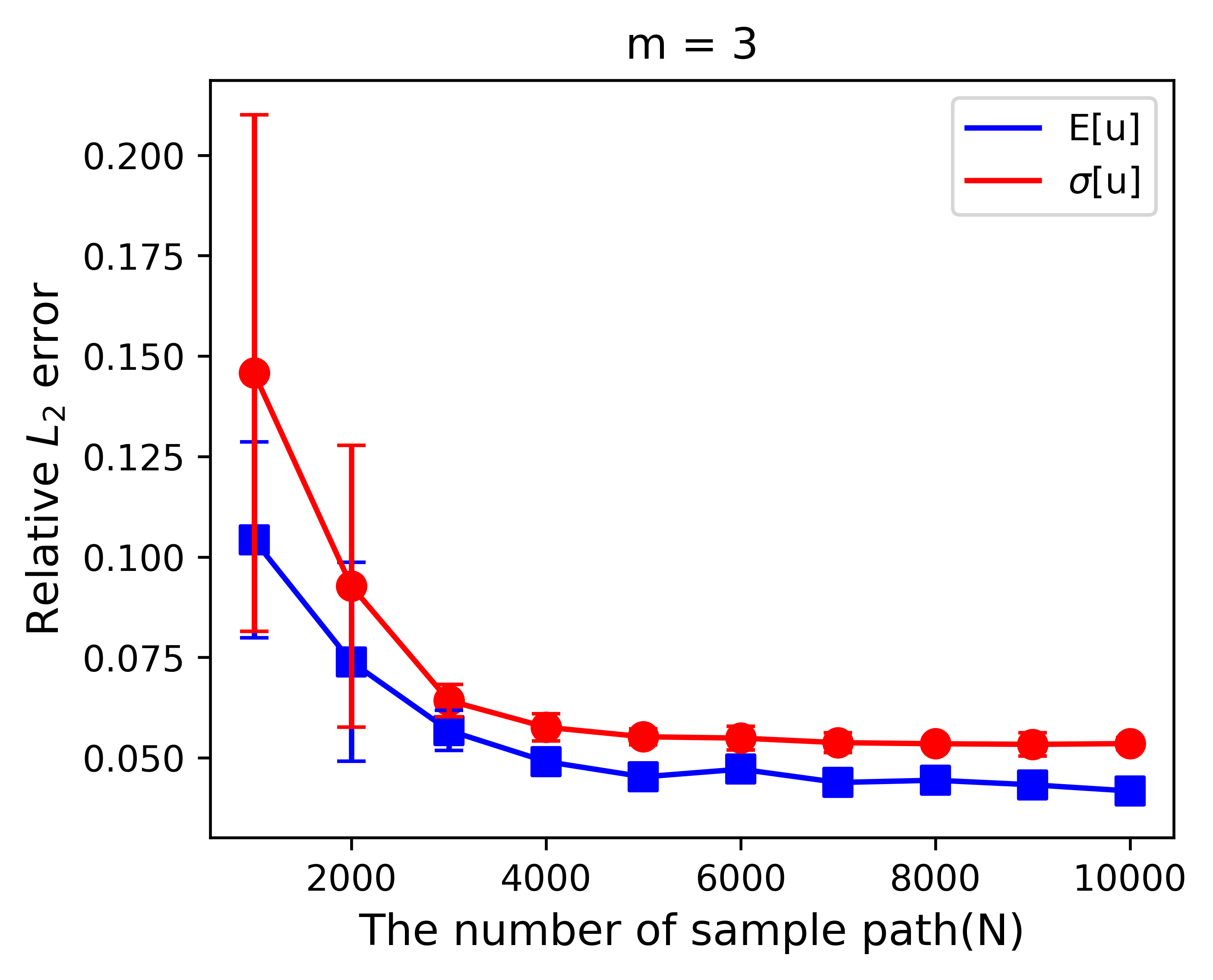}}\hspace{1.0cm} %\qquad
	{\includegraphics[height=0.30\textheight, width=0.46\textwidth]{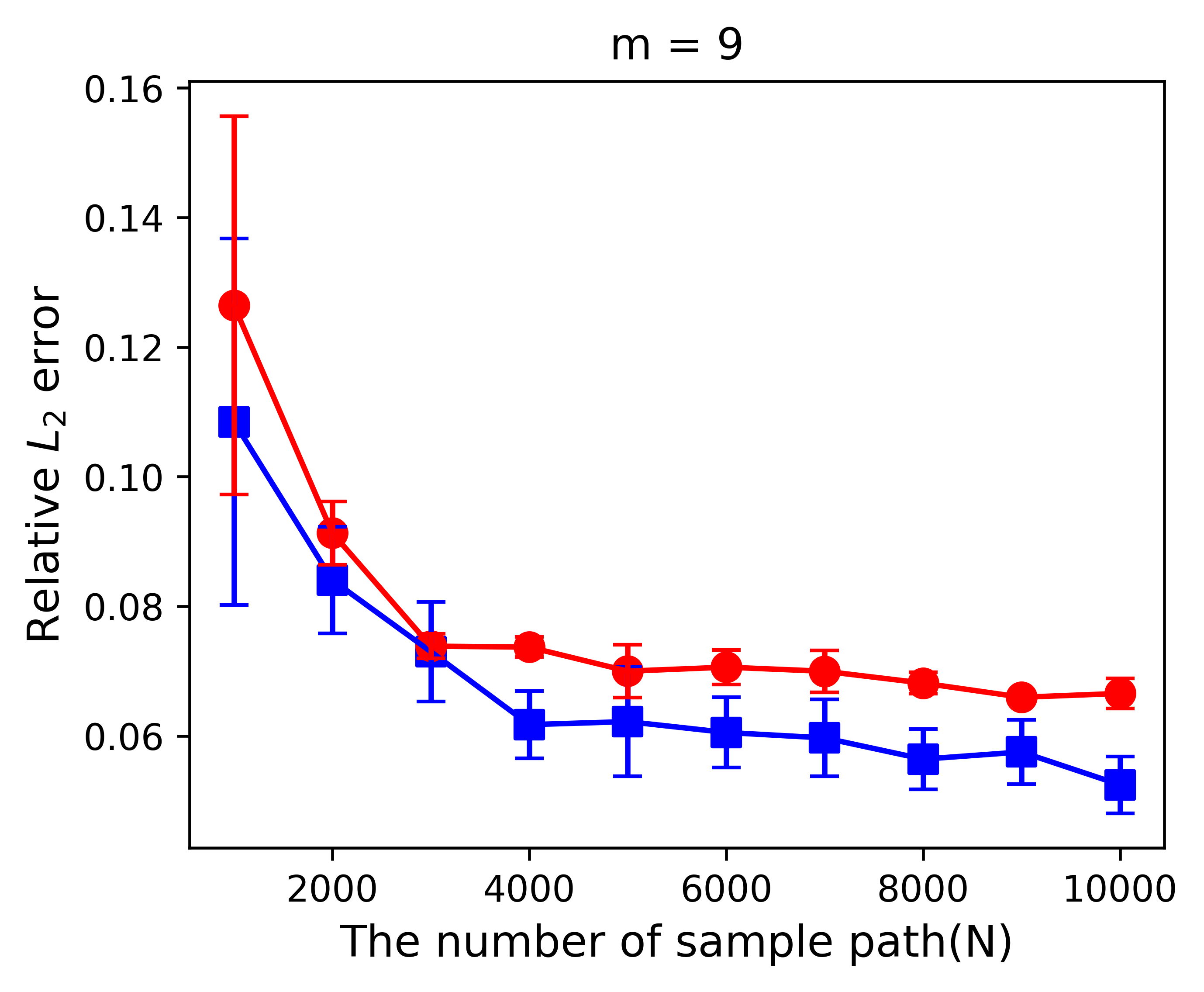}} \\
	\caption{\textbf{1d Stochastic Differential Equation \eqref{eq:1d-sto}.}  Relative $L_2$ errors of the mean and standard deviation between UQ-SONet predictions and reference values, computed from three independent experiments, for $m=3$ (left) and $m=9$ (right).}\label{fig:1d-sde-compare}
\end{figure}

We first investigate the influence of the training dataset size on the performance of UQ-SONet, under the setting where the number of sensors is fixed but their locations are allowed to vary. The latent dimension is fixed at 10, and evaluations are performed for sensor configurations of 3 and 9. Fig.~\ref{fig:1d-sde-compare} reports the relative $L_2$ errors of the predicted mean and standard deviation between different sizes of training datasets, with error bars indicating the results of five independent trials. As the number of training samples increases, the accuracy of both the mean and standard deviation improves steadily and eventually reaches a plateau. In the subsequent experiments, we therefore use 10,000 training samples.

\begin{figure}[!ht]
	\centering
	{\includegraphics[height=0.40\textheight, width=1.0\textwidth]{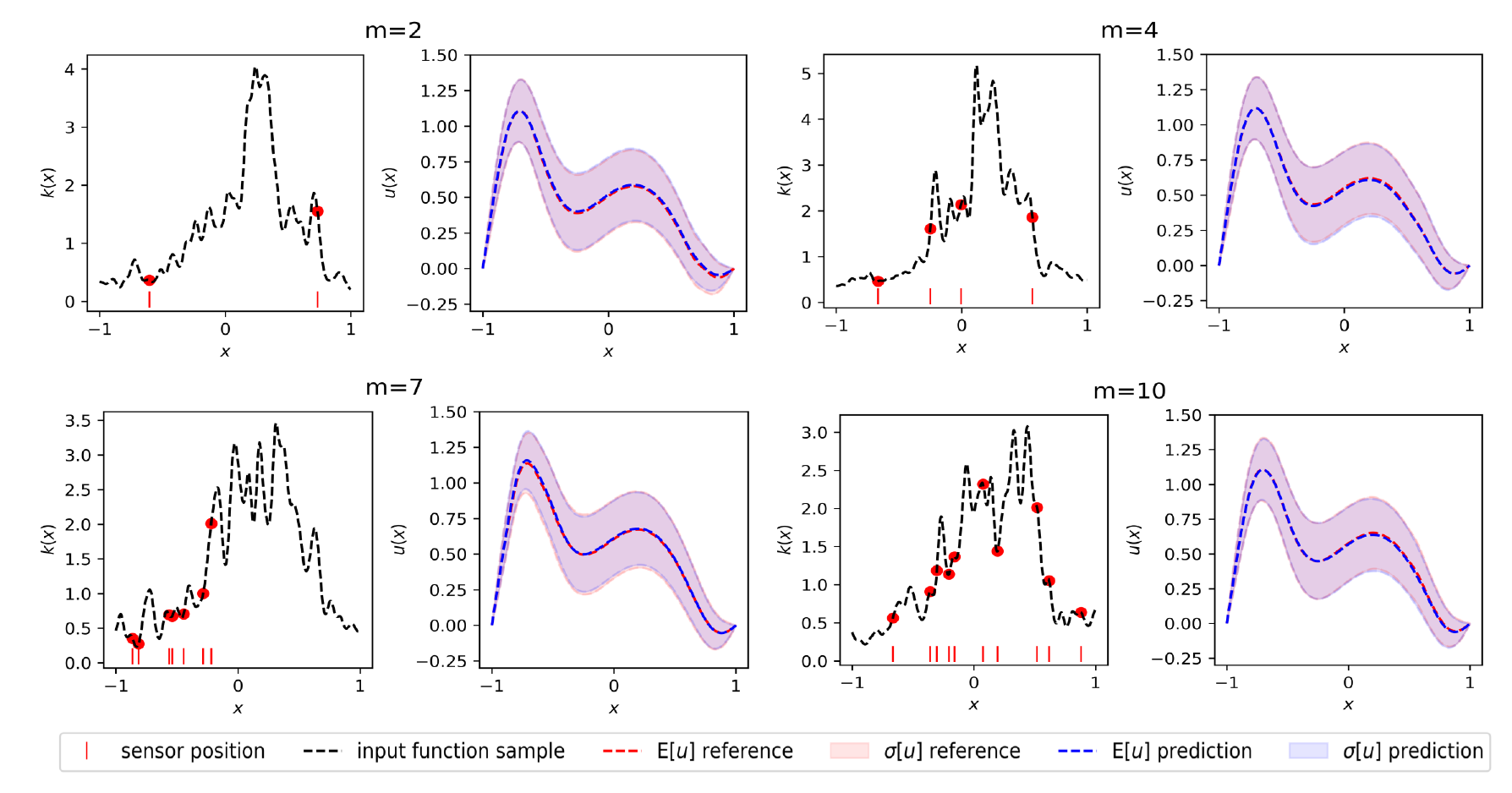}}\\
	\caption{\textbf{1d Stochastic Differential Equation \eqref{eq:1d-sto}.} Reference solution versus UQ-SONet prediction for a representative example with $m=2$ (top left), $m=4$ (top right), $m=7$ (bottom left), $m=10$ (bottom right). Black dashed lines indicate sampled input values, while red markers denote sensor locations. Red and blue dashed lines with shaded regions denote the mean and standard deviation of the reference and predicted solutions, respectively. }\label{fig:1d-sde-plot}
\end{figure}

We further investigate the case where both the number and the locations of the sensors vary, with the number of sensors taken from the set $\mathcal{M} = \{1, 2, 3, 4, 5, 6, 7, 8, 9, 10\}$. In Fig. \ref{fig:1d-sde-plot}, we present the predicted mean and standard deviation for a representative solution, with the number of input function sensors set to $2$, $4$, $7$, and $10$, respectively. The results show that the UQ-SONet predictions align closely with the reference solution. Notably, the solution $u$ retains a degree of uncertainty even when the number of sensors is large. This persistence arises from the two stochastic sources embedded in the operator: while the uncertainty due to insufficient input information diminishes as sensor counts increase, the intrinsic randomness of the operator continues to manifest in the solution. Table \ref{tab:1d-sde-error} reports the relative $L_2$ errors the mean and standard deviation, averaged over five independent experiments. The numerical results indicate that UQ-SONet can consistently recover the conditional distribution of each individual output sample, even in cases where the operator itself exhibits inherent randomness.

\begin{table}[!ht]
	\small
	\centering
	\setlength{\tabcolsep}{8.5pt}
	\renewcommand\arraystretch{1.6}
	\caption{\textbf{1d Elliptic Stochastic Differential Equation \eqref{eq:1d-sto}.} Relative $L_2$ errors of mean and standard deviation between the reference distribution and the UQ-SONet's conditional distribution, where the $\pm$ is calculated from $5$ independent experiments. All values reported are scaled by a factor of $1 \times 10^{-2}$.}
	\label{tab:1d-sde-error}
	\begin{tabular}{c|cccccccc}
		\hline
		$m$ & 1 & 2 & 3 & 4 & 5  \\
		\hline
		$err_{E[u]}$ & $1.76 \pm 0.32$ & $1.79 \pm 0.27$ & $2.29 \pm 0.27$ & $2.04 \pm 0.19$ & $2.37 \pm 0.42$ \\ 
		$err_{\sigma[u]}$ & $3.94 \pm 0.40$ & $3.90 \pm 0.12$ & $4.54 \pm 0.75$ & $3.61 \pm 0.68$ & $5.12 \pm 0.64$  \\ 
		\hline
		$m$ & 6 & 7 & 8 & 9 & 10 \\
		\hline
		$err_{E[u]}$ & $2.16 \pm 0.21$ & $2.38 \pm 0.18$ & $2.45 \pm 0.23$ & $2.34 \pm 0.28$ & $2.40 \pm 0.44$ \\ 
		$err_{\sigma[u]}$ & $4.58 \pm 0.31$ & $5.54 \pm 0.68$ & $4.61 \pm 0.52$ & $5.66 \pm 0.48$ & $4.58 \pm 0.38$ \\ 
		\hline
	\end{tabular}
\end{table}

\subsection{Two-dimensional Stochastic Differential Equation.}\label{subsec:2d-sde}

Next, we consider the two-dimensional elliptic stochastic differential equation:
\begin{equation}\label{eqn:2d-sde}
	\begin{aligned}
			-\nabla \cdot(k(x, y; \omega) \nabla u(x, y; \omega)) & = f(x, y; \omega), \quad x \in \mathcal{D}=[0,1]^2, \quad \omega \in \Omega, \\
			u(x, y; \omega) & = 0, \quad x \in \partial \mathcal{D}.
		\end{aligned}
\end{equation}
both $\log k(x, y; \omega)$ and $f(x, y; \omega)$  are modeled as two-dimensional Gaussian processes with  covariance given by \eqref{eqn:2d-gp}. The diffusion coefficient $\log k(x, y; \omega)$ has zero mean and isotropic correlation lengths $l_1=l_2=0.1$. In contrast, the source term $f(x, y; \omega)$ has a non-zero mean given by $4(sin(2 \pi x) + sin(2 \pi y))$ and the same correlation lengths $l_1=l_2=0.1$.

We now focus on the two-dimensional stochastic operator mapping $\mathcal{G}: f(x, y; \omega) \longmapsto u(x, y; \omega)$. Similar to Section \ref{subsec:2d-pde}, we define the set of sensor counts as $\mathcal{M} = \{1, 2, 3, 4\}$. The sizes of the training and testing datasets, as well as the selection method for the input function sensors, are kept the same as in the previous setting. For each realization of the stochastic processes $k(x, y; \omega)$ and $f(x, y; \omega)$, the corresponding output function is computed using a second-order central difference scheme on a uniform $101 \times 101$ grid. During training, observations are sampled on a uniform $51 \times 51$ grid, while predictions are performed on a finer $101 \times 101$ uniform grid. Four batches of training datasets are pre-generated, and one batch is randomly selected at each iteration. The model is trained for a total of $50,000$ steps. 

\begin{figure}[!ht]
	\centering
	{\includegraphics[height=0.34\textheight, width=1.0\textwidth]{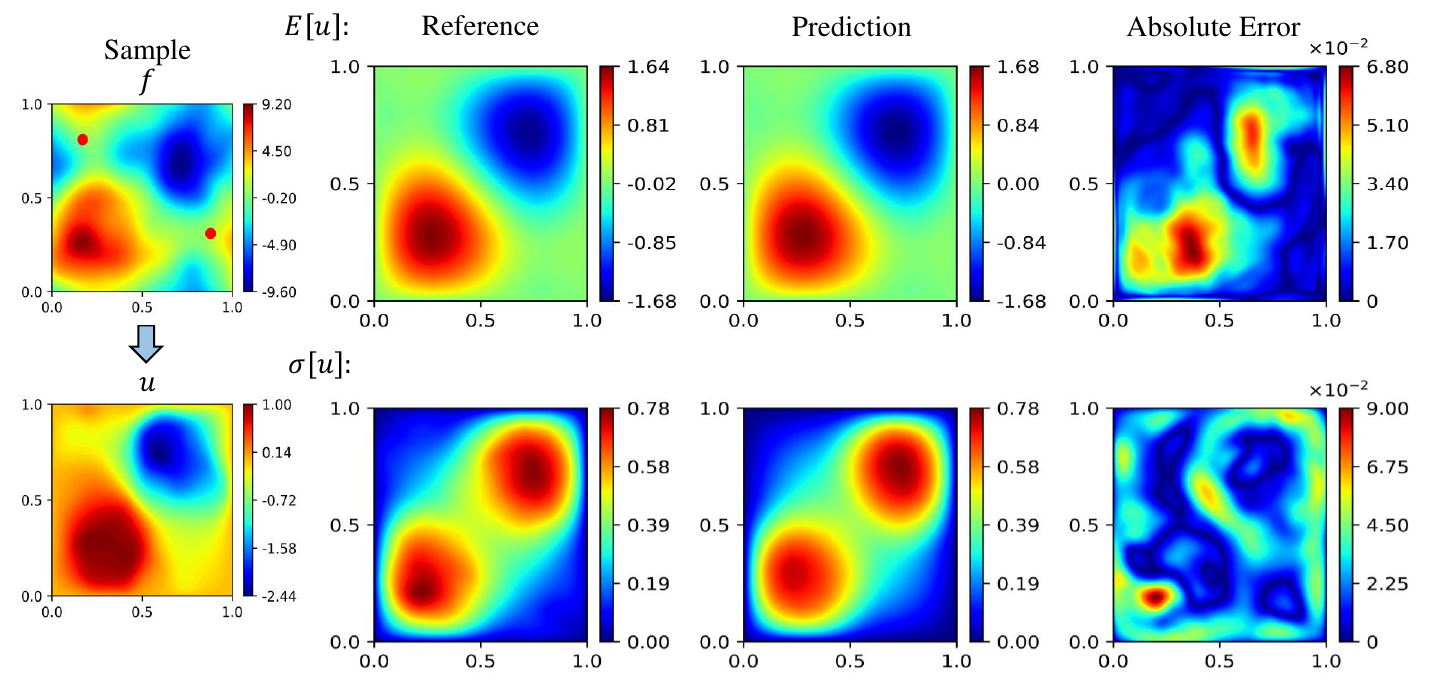}}\\
	\caption{\textbf{2d Stochastic Differential Equation \eqref{eqn:2d-sde}.} \textbf{Left:} Reference samples of the two-dimensional operator map. The red markers indicate sensor locations; \textbf{Top:} the reference mean, the predicted mean by UQ-SONet, and the corresponding absolute error; \textbf{Bottom:} the reference standard deviation, the predicted standard deviation by UQ-SONet, and the corresponding absolute error. }\label{fig:2d-sde-plot}
\end{figure}

The predicted conditional distribution and the reference solution for a representative sample with two input function sensors are shown in Fig. \ref{fig:2d-sde-plot}. Table \ref{tab:2d-sde-error} provides the relative $L_2$ errors of the mean and standard deviation, computed from three independent experiments, evaluated at $x=0.5,x=0.7$ and $y=0.5,y=0.7$. The results demonstrate that UQ-SONet achieves good predictive accuracy for the conditional distribution of individual output samples, even when uncertainty arises from both limited input observations and the intrinsic randomness of the operator.

\begin{table}[!ht]
	\centering
	\small
	\setlength{\tabcolsep}{7.0pt} 
	\renewcommand{\arraystretch}{1.45} 
	\caption{\textbf{2d Stochastic Differential Equation \eqref{eqn:2d-sde}.} Relative $L_2$ errors of the mean and standard deviation between reference distribution and conditional distribution at $x=0.5,x=0.7$ and $y=0.5,y=0.7$ predicted via the UQ-SONet , where $\pm$ indicates the variability across three independent experiments, with all values scaled by a factor of $1 \times 10^{-2}$.}
	\label{tab:2d-sde-error}
	\begin{tabular}{ c | c c | c c | c c | c c }
		\hline
		\multirow{2}{*}{\centering \makecell{$m$}} & \multicolumn{2}{c}{\makecell{$\mathbf{x = 0.5}$}} & \multicolumn{2}{c}{\centering \makecell{$\mathbf{y = 0.5}$}} & \multicolumn{2}{c}{\centering \makecell{$\mathbf{x = 0.7}$}} & \multicolumn{2}{c}{\centering \makecell{$\mathbf{y = 0.7}$}}\\ 
		\cline{2-9}
		& $err_{E[u]}$ & $err_{\sigma[u]}$ & $err_{E[u]}$ & $err_{\sigma[u]}$ & $err_{E[u]}$ & $err_{\sigma[u]}$ & $err_{E[u]}$ & $err_{\sigma[u]}$\\ 
		\hline
		1  & $4.23 \pm 0.43$ & $8.09 \pm 0.14$ & $3.91 \pm 0.16$ & $6.92 \pm 0.45$ & $3.23 \pm 0.38$ & $8.11 \pm 0.64$ & $3.23 \pm 0.33$ & $7.61 \pm 0.58$\\ 
		2  & $7.57 \pm 0.49$ & $8.13 \pm 0.36$ & $5.98 \pm 0.19$ & $7.83 \pm 0.67$ & $4.85 \pm 0.23$ & $8.25 \pm 0.47$ & $4.44 \pm 0.52$ & $6.95 \pm 0.38$\\ 
		3  & $6.92 \pm 0.34$ & $7.26 \pm 1.02$ & $7.45 \pm 0.86$ & $7.57 \pm 0.53$ & $4.72 \pm 0.44$ & $6.52 \pm 0.32$ & $4.25 \pm 0.18$ & $6.96 \pm 0.33$\\ 
		4  & $7.13 \pm 0.59$ & $8.50 \pm 0.18$ & $6.47 \pm 0.33$ & $8.13 \pm 0.19$ & $4.51 \pm 0.43$ & $7.02 \pm 0.34$ & $4.21 \pm 0.32$ & $7.48 \pm 0.10$\\ 
		\hline
	\end{tabular}
\end{table}

\subsection{Navier-Stokes Equation.}\label{subsec:2d-ns}

Finally, we consider the 2D incompressible Navier-Stokes equations in vorticity-velocity form with periodic boundary conditions \cite{li2020fourier}:
\begin{equation}\label{eqn:2d-ns}
	\begin{aligned}
		\partial_t w\left(\mathbf{x}, t\right)+u\left(\mathbf{x}, t\right) \cdot \nabla w\left(\mathbf{x}, t\right) & =\nu \Delta w\left(\mathbf{x}, t\right)+f\left(\mathbf{x}\right), & & \mathbf{x} = \left(x, y\right) \in \left[0,1\right]^2, t \in \left[0, T\right] \\
		\nabla \cdot u\left(\mathbf{x}, t\right) & =0, & & \mathbf{x} = \left(x, y\right) \in \left[0,1\right]^2, t \in \left[0, T\right] \\
		w\left(\mathbf{x}, 0\right) & = w_0\left(\mathbf{x}\right), & & \mathbf{x}  = \left(x, y\right) \in \left[0,1\right]^2
	\end{aligned}
\end{equation}
where $w\left(\mathbf{x}, t\right)$ denotes the vorticity field, defined as $w = \nabla \times u$ and $u\left(\mathbf{x}, t\right)$ is the divergence-free velocity field. The viscosity coefficient is set to $v = 0.001$ in our experiments, with the forcing term given by $f(x,y) = 0.1\sin(2\pi(x+y)) + 0.1\cos(2\pi(x+y))$. We are interested in learning the operator mapping from the initial vorticity field $w_0$ to the vorticity field at target time $T=10$. The initial vorticity field is sampled from a two-dimensional random field $g\left(\mathbf{x}; \omega\right) = x^{1/3}\left(1-x\right)^{1/3}y^{1/3}\left(1-y\right)^{1/3} h\left(\mathbf{x}; \omega\right)$. The random process $h\left(\mathbf{x}; \omega\right)$ is modeled as a zero-mean two-dimensional Gaussian process with covariance defined in \eqref{eqn:2d-gp}, and correlation lengths set to $l_1 = l_2 = 0.1$.

We define the sensor set as $\mathcal{M} = \{1, 2, 3, 4\}$. The sizes of the training and testing datasets, as well as the procedure for selecting sensors for the input functions, follow the setup described in Section~\ref{subsec:2d-pde}. Each realization of the initial condition is discretized on a uniform $100 \times 100$ grid, and the corresponding output function is computed using the stream-function formulation with a pseudospectral method on the same grid. During training, observations are uniformly sampled on a $50 \times 50$ subgrid, whereas predictions are evaluated on the full $100 \times 100$ grid. The training dataset is divided into four batches, each containing $20,000$ samples. At each iteration, one batch is randomly selected for model optimization, and the models are trained for a total of $50,000$ iterations.

\begin{figure}[!ht]
	\centering
	{\includegraphics[height=0.34\textheight, width=1.0\textwidth]{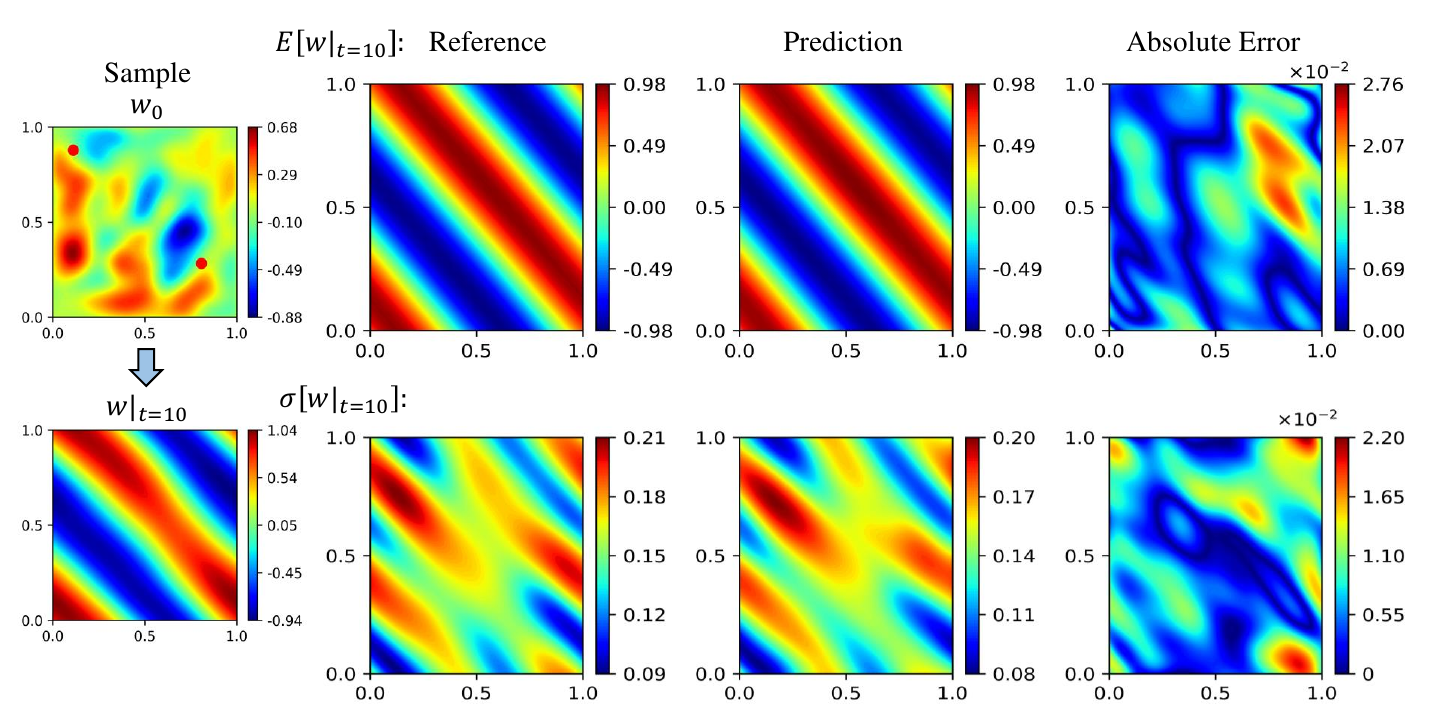}}\\
	\caption{\textbf{Navier-Stokes Equation \eqref{eqn:2d-ns}.} \textbf{Left:} Reference samples of the operator map. The red markers indicate sensor locations; \textbf{Top:} the reference mean, the predicted mean by UQ-SONet, and the corresponding absolute error; \textbf{Bottom:} the reference standard deviation, the predicted standard deviation by UQ-SONet, and the corresponding absolute error. }\label{fig:2d-ns-plot}
\end{figure}

In Fig. \ref{fig:2d-ns-plot}, we present the mean and standard deviation of the conditional distribution predicted by UQ-SONet for a representative sample with 2 input function sensors, compared against the corresponding reference solution. The relative $L_2$ errors of the predicted mean and standard deviation, averaged over three independent runs, are reported in Table \ref{tab:2d-ns-error}, together with a comparison against the deterministic VIDON method. The results highlight both the precision and robustness of the proposed UQ-SONet framework.

\begin{table}[!ht]
	\centering
	\small
	\setlength{\tabcolsep}{13pt} 
	\renewcommand{\arraystretch}{1.65} 
	\caption{\textbf{Navier-Stokes Equation \eqref{eqn:2d-ns}.} The relative $L_2$ errors of the mean and standard deviation between the reference distribution and the predicted conditional distribution obtained by UQ-SONet method (with comparison to VIDON, which provides only mean estimates), where the $\pm$ is calculated from $3$ independent experiments. All values reported  are scaled by a factor of $1 \times 10^{-2}$.}
	\label{tab:2d-ns-error}
	\begin{tabular}{ c | c c c }
		\hline
		\multirow{2}{*}{\centering \makecell{$m$}} & \textbf{VIDON} & \multicolumn{2}{c}{\textbf{UQ-SONet}} \\ 
		\cline{2-4}
		& $err_{E[w|_{t=10}]}$ & $err_{E[w|_{t=10}]}$ & $err_{\sigma[w|_{t=10}]}$  \\ 
		\hline
		1  & $6.01 \pm 0.64$ & $2.35 \pm 1.03$ & $7.32 \pm 1.08$ \\ 
		2  & $5.98 \pm 0.36$ & $1.89 \pm 0.60$ & $7.02 \pm 0.78$ \\ 
		3  & $8.74 \pm 0.78$ & $2.55 \pm 0.46$ & $6.25 \pm 0.54$ \\ 
		4  & $8.08 \pm 0.75$ & $2.69 \pm 0.38$ & $6.52 \pm 0.25$ \\ 
		\hline
	\end{tabular}
\end{table}

\section{Conclusion}
We have proposed UQ-SONet, a reliable operator learning framework with built-in UQ. The model integrates a set transformer embedding within a cVAE, achieving permutation invariance with respect to both the number and locations of input sensors and thereby enhancing expressive power. The framework has been applied to a wide range of PDE and SDE problems with random inputs. For deterministic operators, UQ-SONet effectively quantifies output uncertainty arising from incomplete input observations. The method can also be naturally extended to scenarios with sparse and noisy observations, as well as cases where the operator itself exhibits intrinsic randomness. Future work will focus on developing active learning strategies, where uncertainty estimates from UQ-SONet guide the adaptive placement of sensors to maximize information gain, as well as on studying long-term rollouts in time-dependent systems, where accumulating uncertainty poses significant challenges. In addition, the current framework uses a conditionally Gaussian decoder, which may be restrictive for strongly non-Gaussian or multimodal posteriors. Extending the model to more expressive latent or decoder distributions, together with alternative generative or efficient sampling approaches such as Hamiltonian Monte Carlo (HMC) and diffusion models, may improve the approximation of the target distribution. It is also of interest to extend the framework to settings with limited output observations, where uncertainty arises from both sparse input data and insufficient output measurements.

\section*{Acknowledgments}
The first and second authors are supported by the NSF of China (No. 92270115 and 12071301). The third author is supported by the NSF of China (No. 12571463). The last author is supported by the NSF of China (under grants 12288201 and 12461160275), and the high-level talent research start-up project funding of Henan Academy of Sciences (No. 232019024), and the science challenge project (No. TZ2025006).

\FloatBarrier

\bibliography{references-first}{}
%\printbibliography

% \bibliographystyle{plainnat}  
%\bibliographystyle{plainnat}  
%\bibliographystyle{siam}
%\bibliographystyle{elsarticle-num}
\newpage
\appendix
\setcounter{remark}{0}
\setcounter{theorem}{0}
\setcounter{definition}{0}
\setcounter{figure}{0}
\setcounter{table}{0}
\renewcommand{\thetheorem}{\Alph{section}.\arabic{theorem}}
\renewcommand{\thedefinition}{\Alph{section}.\arabic{definition}}
\renewcommand{\theremark}{\Alph{section}.\arabic{remark}}
\renewcommand{\thefigure}{\Alph{section}.\arabic{figure}}
\renewcommand{\thetable}{\Alph{section}.\arabic{table}}

\section{Appendix A}

\subsection{Hyperparameters and neural network architectures used in the numerical examples}\label{appendix:network}

In this appendix, we summarize the hyperparameters and neural network architectures employed in all numerical experiments, as listed in Table~\ref{tab:NN}. The VIDON method adopts the same network structures and hyperparameters as UQ-SONet model.

\begin{table}[!ht]
	%\small
	\centering
	\renewcommand\arraystretch{1.8}
	\setlength{\tabcolsep}{5.2pt}
	\caption{The sizes of $\Psi_x$, $\Psi_{\kappa}$, $\{w_l\}_{1 \leq l \leq H}$,  $\{v_l\}_{1 \leq l \leq H}$, $B$, $T$, and $\mathcal{E}$ are reported, together with key model parameters, including the variance of the artificial noise in the output function, the dimensions of the embedding and latent variable, and the learning rate. Here, $2 \times 40$ denotes a network with two hidden layers, each comprising $40$ neurons.}
	\label{tab:NN}
	\begin{tabular}{c|ccccccccc|c|cc }
		\hline
		&  $\Lambda_x$  &  $\Lambda_{\kappa}$  &  $\{w_l\}_{1 \leq l \leq H}$  &  $\{v_l\}_{1 \leq l \leq H}$ & $B$ & $T$ & $\mathcal{E}$ & $d_{z}$ & $d_{emb}$ & $\sigma_u^2$ &  $lr$\\
		\hline
		1d PDE   & 2 $\times$ 40 & 2 $\times$ 40 & 4 $\times$ 32 & 4 $\times$ 32 & 4 $\times$ 64 &  4 $\times$ 64 &  4 $\times$ 64 & $10$ & 2 &  $10^{-3}$ & \multirow{3}{*}{\centering $10^{-4}$} \\
		\cline{1-11}
		1d SDE   & 2 $\times$ 40 & 2 $\times$ 40 & 4 $\times$ 32 & 4 $\times$ 32 & 4 $\times$ 64 &  4 $\times$ 64 &  4 $\times$ 64 & $10$ & 2 & \multirow{2}{*}{\centering $10^{-4}$} &  \\
		\cline{1-10}
		\makecell{2d \\ PDE/SDE/NS}  & 4 $\times$ 40 & 4 $\times$ 40 & 4 $\times$ 64 & 4 $\times$ 64 & 4 $\times$ 128 &  4 $\times$ 128 &  4 $\times$ 128 & $100$ & 3 & & \\  
		\hline
	\end{tabular}
\end{table}

\subsection{Computation of Reference Solutions}\label{appendix:reference}

For completeness, we briefly describe how the reference distributions of operator outputs are obtained in the numerical examples presented in this work.  
Assume that the prior distribution of the input function $\kappa$ is a Gaussian process with mean function $\mu$ and covariance kernel $\mathcal{K}$.  
Given observations $y_i = \kappa(x_i)$ at sensor locations $x_i$ ($i=1,\ldots,m$) from a particular realization of $\kappa$, the posterior of $\kappa$ remains a Gaussian process whose mean function and covariance kernel can be expressed as:
\begin{equation}
	\begin{aligned}
		\mu_{\text{post}}\left(x\right) = \mu\left(x\right) + \mathcal{K}\left(x, x_{\text{obs}}\right) \mathcal{K}\left(x_{\text{obs}}, x_{\text{obs}}\right)^{-1} \left(y_{\text{obs}} - \mu\left(x_{\text{obs}}\right)\right),\\
		\mathcal{K}_{\text{post}}\left(x, x'\right) = \mathcal{K}\left(x, x'\right) - \mathcal{K}\left(x, x_{\text{obs}}\right) \mathcal{K}\left(x_{\text{obs}}, x_{\text{obs}}\right)^{-1} \mathcal{K}\left(x_{\text{obs}}, x'\right).
	\end{aligned}
\end{equation}
where $x_{\text{obs}}=[x_1,\ldots,x_m]^{T}$ and $y_{\text{obs}}=[y_1,\ldots,y_m]^{T}$. We can then draw exact samples from the posterior of $\kappa$ and compute the corresponding output function $u$ using a standard finite difference method.  
In the numerical experiments, for each observation set we generate $1{,}000$ posterior samples in order to evaluate the UQ results.

Note that in Sections \ref{subsec:1d-pde}, \ref{subsec:1d-sde} and \ref{subsec:2d-sde}, the operator models take observations of $k$ as input, whereas the reference solutions are computed using $\log k$ as the input function, since $\log k$ follows a Gaussian process.

Furthermore, when the observations of $\kappa$ are corrupted by additive Gaussian noise with distribution $\mathcal{N}(0,\sigma^2)$, the posterior distribution of $\kappa$ can be written as $\mathcal{GP}\left(\hat{\mu}_{\text{post}}, \hat{\mathcal{K}}_{\text{post}}\right)$, where
\begin{equation}
	\begin{aligned}
		\hat{\mu}_{\text{post}}\left(x\right) = \mu\left(x\right) + \mathcal{K}\left(x, x_{\text{obs}}\right) \left[ \mathcal{K}\left(x_{\text{obs}}, x_{\text{obs}}\right) + \sigma^2 I \right]^{-1} \left(\hat{y}_{\text{obs}} - \mu\left(x_{\text{obs}}\right)\right),\\
		\hat{\mathcal{K}}_{\text{post}}\left(x, x'\right) = \mathcal{K}\left(x, x'\right) - \mathcal{K}\left(x, x_{\text{obs}}\right) \left[ \mathcal{K}\left(x_{\text{obs}}, x_{\text{obs}}\right) + \sigma^2 I \right]^{-1} \mathcal{K}\left(x_{\text{obs}}, x'\right).
	\end{aligned}
\end{equation}
where $\hat{y}_{\text{obs}}=[\hat{y}_1, \ldots,\hat{y}_m]^{T}$ denotes the noisy observations. The computation of the corresponding reference solutions follows in a similar manner.  

\subsection{Derivation of the UQ-SONet loss from functional VAE principles}\label{appendix:messure}

Based on the theoretical framework of~\cite{seidman2023variational}, we derive a training loss function that does not rely on the discretization of the output function \( u \). Here, we assume that \( u \) is defined on a bounded domain \( \Omega_u \subset \mathbb{R}^{d_y} \) with finite measure. 

We define the decoder in function space as
\[
u_\sigma = \frac{u}{\sigma} = \frac{\hat{\mathcal{G}}(h(\myO), z)}{\sigma} + \eta,
\]
where \( \sigma > 0 \) is a hyperparameter, \( \eta \sim \mathbb{V} \), and \( \mathbb{V} \) is a Gaussian measure on \( \Omega_u \).
Accordingly, the encoder is defined as
\[
z \mid \myO, u \sim \mathcal{N} \left( z \mid \mu_z(h(\myO), u), \Sigma_z(h(\myO), u) \right).
\]
In what follows, to be consistent with the notation in~\cite{seidman2023variational}, we denote by \( \mathbb{P} \) and \( \mathbb{Q} \) the probability measures defined by the decoder and encoder, respectively.

By considering \( u_\sigma \) as the output function, and according to Theorem~4.1 and (9) in~\cite{seidman2023variational}, for given observations \( \myO \), we obtain:
\begin{eqnarray*}
\mathcal{L}_{f}(o) & \triangleq & \mathbb{E}_{u_{\sigma}|\scalebox{0.6}{$\mathcal{O}$}\sim\mathbb{P}_{u_{\sigma}|\scalebox{0.4}{$\mathcal{O}$}},z|u_{\sigma},\scalebox{0.6}{$\mathcal{O}$}\sim\mathbb{Q}_{z|u_{\sigma},\scalebox{0.4}{$\mathcal{O}$}}}\left[\frac{1}{2}\left\Vert \frac{\hat{\mathcal{G}}\left(h(\myO),z\right)}{\sigma}\right\Vert _{L_{2}}^{2}-\left\langle \frac{\hat{\mathcal{G}}\left(h(\myO),z\right)}{\sigma},u_{\sigma}\right\rangle ^{\sim}\right]\\
 &  & +\mathbb{E}_{u_{\sigma}|o\sim\mathbb{P}_{u_{\sigma}|\scalebox{0.4}{$\mathcal{O}$}}}\left[\mathrm{KL}\left(\mathbb{Q}_{z|u_{\sigma}}||\mathcal{N}\left(z\mid0,I\right)\right)\right]\\
 & \ge & -\mathbb{E}_{u_{\sigma}|\scalebox{0.6}{$\mathcal{O}$}\sim\mathbb{P}_{u_{\sigma}|\scalebox{0.4}{$\mathcal{O}$}}}\left[\log\frac{\mathrm{d}\mathbb{P}_{u_{\sigma}}}{\mathrm{d}\mathbb{V}}\right]\\
 & = & -\sigma\mathbb{E}_{u|\scalebox{0.6}{$\mathcal{O}$}\sim\mathbb{P}_{u|\scalebox{0.4}{$\mathcal{O}$}}}\left[\log\frac{\mathrm{d}\mathbb{P}_{u}}{\mathrm{d}\mathbb{V}}\right],
\end{eqnarray*}
where can be interpreted as a generalization of the inner product on \( L^2(\Omega_u) \),  which in this case reduces to \( \left\langle D(z), u \right\rangle \).

Therefore, maximum likelihood estimation of the operator can be performed by minimizing \( \mathbb{E}_{\scalebox{0.6}{$\mathcal{O}$} \sim \mathrm{data}} [\mathcal{L}_f(\myO)] \).  
Based on the definitions of the decoder and encoder given above, we have
\begin{eqnarray}
\mathcal{L}_{f} & \triangleq & \mathbb{E}_{\scalebox{0.6}{$\mathcal{O}$}\sim\mathrm{data}}\left[\mathcal{L}_{f}(\myO)\right]\nonumber \\
 & = & \mathbb{E}_{u_{\sigma},\scalebox{0.6}{$\mathcal{O}$}\sim\mathrm{data}}\left[\mathrm{KL}\left(\mathbb{Q}_{z\mid u_{\sigma},\scalebox{0.6}{$\mathcal{O}$}}\,\|\,\mathcal{N}(z\mid0,I)\right)\right]\nonumber \\
 &  & +\frac{1}{2}\,\mathbb{E}_{u_{\sigma},\scalebox{0.6}{$\mathcal{O}$}\sim\mathrm{data},z\mid u_{\sigma},\scalebox{0.6}{$\mathcal{O}$}\sim\mathbb{Q}_{z\mid u_{\sigma}}}\left[\left\Vert \frac{\hat{\mathcal{G}}(h(\myO),z)}{\sigma}-u_{\sigma}\right\Vert _{L_{2}}^{2}\right]+\mathrm{const}\nonumber \\
 & = & \mathbb{E}_{u,\scalebox{0.6}{$\mathcal{O}$}\sim\mathrm{data}}\left[\mathrm{KL}\left(\mathbb{Q}_{z\mid u}\,\|\,\mathcal{N}(z\mid0,I)\right)\right]\nonumber \\
 &  & +\frac{1}{2\sigma^{2}}\,\mathbb{E}_{u,\scalebox{0.6}{$\mathcal{O}$}\sim\mathrm{data},z\mid u,\scalebox{0.6}{$\mathcal{O}$}\sim\mathbb{Q}_{z\mid u,\scalebox{0.4}{$\mathcal{O}$}}}\left[\left\Vert \hat{\mathcal{G}}(h(\myO),z)-u\right\Vert _{L_{2}}^{2}\right]+\mathrm{const}\nonumber \\
 & = & \frac{1}{2}\,\mathbb{E}_{u,\scalebox{0.6}{$\mathcal{O}$}\sim\mathrm{data}}\left[-\log|\Sigma_{z}(h(\myO),u)|+\mathrm{tr}(\Sigma_{z}(h(\myO),u))+\|\mu_{z}(h(\myO),u)\|^{2}-d_{z}\right]\nonumber \\
 &  & +\frac{1}{2\sigma^{2}}\,\mathbb{E}_{u,\scalebox{0.6}{$\mathcal{O}$}\sim\mathrm{data},z\mid u,\scalebox{0.6}{$\mathcal{O}$}\sim\mathbb{Q}_{z\mid u,\scalebox{0.4}{$\mathcal{O}$}}}\left[\left\Vert \hat{\mathcal{G}}(h(\myO),z)-u\right\Vert _{L_{2}}^{2}\right]+\mathrm{const},\label{eq:functional-loss}
\end{eqnarray}
where \( \mathrm{const} \) denotes the term independent of model parameters.

It should be noted that for the discretized representation \( \bar{u} \) of \( u \) discussed in the main text, when the grid points are uniformly distributed over \( \Omega_u \) and the number of points is sufficiently large, we have
\[
\sum_{i=1}^{M} \frac{ \left( \hat{\mathcal{G}}(h(\myO), z)(y_i) - u(y_i) \right)^2 }{M} \approx \frac{ \left\| \hat{\mathcal{G}}(h(\myO), z) - u \right\|_{L_2}^2 }{|\Omega_u|}.
\]
Based on the above relation, by comparing \eqref{eq:functional-loss} with the loss function \( \mathcal{L} \) defined in \eqref{eq:loss}, it can be seen that if we choose \( \sigma=\sqrt{|\Omega_{u}|}\sigma_{u} \), then \( \mathcal{L} \) can be interpreted as a discretized approximation of \( \mathcal{L}_{f} \).

\subsection{Deep Ensemble Baseline Based on VIDON}\label{appendix:deepensem}
To further assess empirical uncertainty estimation, we compare UQ-SONet with a deep ensemble baseline built on VIDON. The ensemble is constructed by training VIDON multiple times with different random initializations and data shuffling, and using the ensemble statistics for prediction and uncertainty estimation. This comparison is intended as a supplementary empirical reference. The results show that the deep ensemble provides accurate mean predictions, but its estimated uncertainty bands are consistently narrower than those of the reference solution, indicating an underestimation of the target conditional uncertainty. This suggests that ensemble-based uncertainty mainly reflects variability across trained deterministic predictors, whereas UQ-SONet is explicitly designed to model the conditional distribution $p(u \mid \myO)$ and therefore provides more reliable uncertainty quantification under sparse and variable observations.

\begin{figure}[!ht]
    \centering
    {\includegraphics[height=0.40\textheight, width=1.0\textwidth]{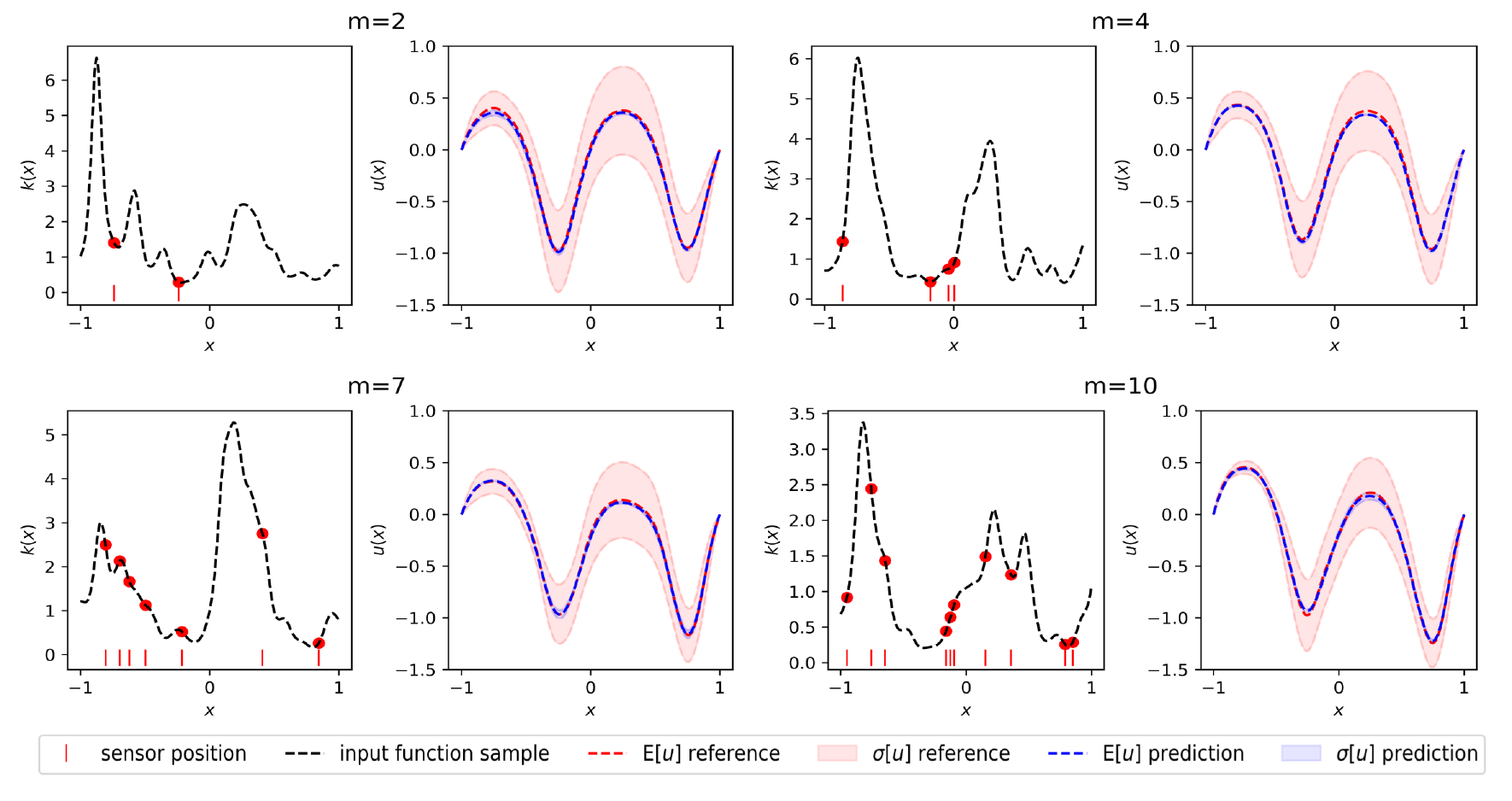}}\\
    \caption{\textbf{Diffusion Equation.} Mean and standard deviation predicted by the deep ensemble based on VIDON, compared with the reference solution. The deep ensemble achieves accurate mean prediction, but its uncertainty bands are consistently narrower than the reference ones.}\label{fig:1d-pde-ensemble-plot}
\end{figure}

\subsection{Summary of Computational Costs}\label{appendix:cost}
In this appendix, we provide a quantitative assessment of the computational cost of all numerical experiments. The corresponding results are summarized in Table \ref{tab:training-cost}, including the training time and inference time. To ensure consistency and fairness in the comparison, all timing results reported in Table \ref{tab:training-cost} were obtained on an NVIDIA GeForce RTX 3090 GPU. As shown in Table \ref{tab:training-cost}, under identical experimental settings, UQ-SONet incurs higher computational cost than VIDON in both the training and inference stages. This is because UQ-SONet introduces an additional encoder network during training, while in the inference stage it also requires latent variable sampling and the computation of predictive statistics for uncertainty quantification. Nevertheless, this additional cost is well justified, as it enables the model to characterize the conditional distribution in a more comprehensive manner.

\begin{table}[!ht]
	\centering
	\small
	\setlength{\tabcolsep}{8pt} 
	\renewcommand{\arraystretch}{1.95} 
	\caption{\textbf{Computational costs.} Training time (hours) and inference time (seconds) are reported for each experimental setting.}
	\label{tab:training-cost}
	\begin{tabular}{ c | c c | c c | c | c | c c c}
		\hline
		{\centering Section} & \multicolumn{2}{c}{\textbf{4.1}} & \multicolumn{2}{c}{\textbf{4.2}} & \multicolumn{1}{c}{\textbf{4.3}} & \multicolumn{1}{c}{\textbf{4.4}} & \textbf{4.5}\\ 
		\hline
		{\centering Method} & VIDON & UQ-SONet & VIDON & UQ-SONet & UQ-SONet & UQ-SONet & UQ-SONet \\ 
		\hline
		\makecell[c]{Train time (h)}  & $5.24$ & $6.03$ & $4.04$ & $4.37$ & $6.01$ & $11.34$  & $10.32$ \\
		\makecell[c]{Infer time (s)}  & $0.025$ & $0.030$ & $0.012$ & $0.028$ & $0.031$ & $0.029$ & $0.026$ \\ 
		\hline
	\end{tabular}
\end{table}

\end{document}